\documentclass[a4paper,11pt,fleqn]{book}

\usepackage{algorithm}
\usepackage{tikz}
\usepackage{algorithmic}
\usepackage{multirow}
\usepackage{makecell}

\usepackage[T1]{fontenc}
\usepackage[utf8]{inputenc}
\usepackage[french,german,english]{babel}

\usepackage{fourier} 
\usepackage{setspace} 

\usepackage{graphicx,xcolor}
\graphicspath{{images/}}

\usepackage{subfig}
\usepackage{booktabs}
\usepackage{lipsum}
\usepackage{microtype}
\usepackage{url}
\usepackage[final]{pdfpages}

\usepackage{fancyhdr}

\fancypagestyle{plain}{
	\fancyhf{}

	\fancyfoot[R]{\thepage}
	}
\fancypagestyle{addpagenumbersforpdfimports}{
	\fancyhead{}
	
	\fancyfoot{}
    \fancyfoot[R]{\thepage}
}

\usepackage{listings}
\usepackage{hyperref}
\hypersetup{pdfborder={0 0 0},
	colorlinks=true,
	linkcolor=black,
	citecolor=black,
	urlcolor=black}
\makeatletter
\def\cleardoublepage{\clearpage\if@twoside \ifodd\c@page\else
    \hbox{}
    \thispagestyle{empty}
    \newpage
    \if@twocolumn\hbox{}\newpage\fi\fi\fi}
\makeatother \clearpage{\pagestyle{plain}\cleardoublepage}

\usepackage{color}
\usepackage{tikz}
\usepackage[explicit]{titlesec}
\newcommand*\chapterlabel{}
\titleformat{\chapter}[display]  
	{\normalfont\bfseries\Huge} 
	{\gdef\chapterlabel{\thechapter\ }}     
 	{0pt} 
 	  {\begin{tikzpicture}[remember picture,overlay]
    \node[yshift=-8cm] at (current page.north west)
      {\begin{tikzpicture}[remember picture, overlay]
        \draw[fill=black] (0,0) rectangle(35.5mm,15mm);
        \node[anchor=north east,yshift=-7.2cm,xshift=34mm,minimum height=30mm,inner sep=0mm] at (current page.north west)
        {\parbox[top][30mm][t]{15mm}{\raggedleft $\phantom{\textrm{l}}$\color{white}\chapterlabel}};  
        \node[anchor=north west,yshift=-7.2cm,xshift=37mm,text width=\textwidth,minimum height=30mm,inner sep=0mm] at (current page.north west)
              {\parbox[top][30mm][t]{\textwidth}{\color{black}#1}};
       \end{tikzpicture}
      };
   \end{tikzpicture}
   \gdef\chapterlabel{}
  } 

\titlespacing*{\chapter}{0pt}{50pt}{30pt}
\titlespacing*{\section}{0pt}{13.2pt}{*0}  
\titlespacing*{\subsection}{0pt}{13.2pt}{*0}
\titlespacing*{\subsubsection}{0pt}{13.2pt}{*0}

\newcounter{myparts}
\newcommand*\partlabel{}
\titleformat{\part}[display]  
	{\normalfont\bfseries\Huge} 
	{\gdef\partlabel{\thepart\ }}     
 	{0pt} 
 	  {\setlength{\unitlength}{20mm}
	  \addtocounter{myparts}{1}
	  \begin{tikzpicture}[remember picture,overlay]
    \node[anchor=north west,xshift=-65mm,yshift=-6.9cm-\value{myparts}*20mm] at (current page.north east) 
      {\begin{tikzpicture}[remember picture, overlay]
        \draw[fill=black] (0,0) rectangle(62mm,20mm);   
        \node[anchor=north west,yshift=-6.1cm-\value{myparts}*20mm,xshift=-60.5mm,minimum height=30mm,inner sep=0mm] at (current page.north east)
        {\parbox[top][30mm][t]{55mm}{\raggedright \color{white}Part \partlabel $\phantom{\textrm{l}}$}};  
        \node[anchor=north east,yshift=-6.1cm-\value{myparts}*20mm,xshift=-63.5mm,text width=\textwidth,minimum height=30mm,inner sep=0mm] at (current page.north east)
              {\parbox[top][30mm][t]{\textwidth}{\raggedleft \color{black}#1}};
       \end{tikzpicture}
      };
   \end{tikzpicture}
   \gdef\partlabel{}
  } 

\usepackage{amsmath}
\makeatletter
\def\resetMathstrut@{%
  \setbox\z@\hbox{%
    \mathchardef\@tempa\mathcode`\(\relax
      \def\@tempb##1"##2##3{\the\textfont"##3\char"}%
      \expandafter\@tempb\meaning\@tempa \relax
  }%
  \ht\Mathstrutbox@1.2\ht\z@ \dp\Mathstrutbox@1.2\dp\z@
}
\makeatother

\usepackage{subfiles}

\usepackage[version=4]{mhchem}
\usepackage{chemist}
\usepackage{chemfig}

\usepackage{csquotes} 
\usepackage[backend=biber,style=apa]{biblatex} 
\begin{document}
\frontmatter
\begin{titlepage}
\begin{center}
\sffamily

\null\vspace{2cm}
{\huge Towards Robust Drone Vision in the Wild} \\[24pt] 
\textcolor{gray}{\small{Master Thesis}}
    
\vfill

\begin{tabular} {cc}
\parbox{0.3\textwidth}{\includegraphics[width=4cm]{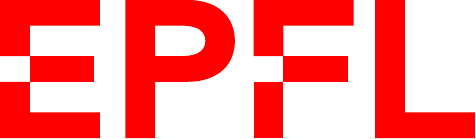}}
&
\parbox{0.7\textwidth}{%
%
    By the student \\ [4pt]
    \null \hspace{3em} Xiaoyu Lin\\[9pt]
%
\small
Approved by the Examining Committee:\\[4pt]
%
    Prof. Süsstrunk Sabine, Thesis Advisor \\
    Prof. Nasrollahi Kamal, External Expert \\
    Dr. Majed El Helou,  Thesis Supervisor \\
%
\\
Image and Visual Representation Lab\\
School of Computer and Communication Sciences\\
Lausanne, EPFL, 2022}
\end{tabular}
\end{center}
\vspace{2cm}
\end{titlepage}

\setcounter{page}{0}

\cleardoublepage
\chapter*{Abstract}
~\newline~\newline~
\addcontentsline{toc}{chapter}{Abstract} 

The past few years have witnessed the burst of drone-based applications where computer vision plays an essential role. However, most public drone-based vision datasets focus on detection and tracking. On the other hand, the performance of most existing image super-resolution methods is sensitive to the dataset, specifically, the degradation model between high-resolution and low-resolution images. In this thesis, we propose the first image super-resolution dataset for drone vision. Image pairs are captured by two cameras on the drone with different focal lengths. We collect data at different altitudes and then propose pre-processing steps to align image pairs. Extensive empirical studies show domain gaps exist among images captured at different altitudes. Meanwhile, the performance of pretrained image super-resolution networks also suffers a drop on our dataset and varies among altitudes. Finally, we propose two methods to build a robust image super-resolution network at different altitudes. The first feeds altitude information into the network through altitude-aware layers. The second uses one-shot learning to quickly adapt the super-resolution model to unknown altitudes. Our results reveal that the proposed methods can efficiently improve the performance of super-resolution networks at varying altitudes.

\vskip0.5cm
Key words: drone vision, image super-resolution, domain adaptation, few-shot learning.




\tableofcontents
\cleardoublepage

\setlength{\parskip}{1em}

\mainmatter
\chapter{Introduction}


\section{Background}

The history of using drones dates back more than one hundred years ago (\cite{akbari2021applications}). Since then, drone technology has endured and been developed mainly for military use. In 2006, DJI, a leader company in the commercial and individual drone industry, developed the first commercial drone. After that, the past few years have witnessed the burst of drones for both commercial and private usage.

Compared with other enduring platforms such as airplanes and satellites, drones can obtain higher-resolution images and achieve better low-altitude maneuverability. It can fly at various speeds indoors or outdoors and control its position, making it an ideal substitute for humans in certain situations, like exploring dangerous places. Moreover, as a platform, the drone can be mounted with multiple sensors and used for different tasks. For example, a drone with sensors to detect its surroundings can automatically avoid obstacles while flying in complicated environments. Most importantly, the drone is much cheaper for maintenance. However, due to the constraint of battery life, the drone can only work continuously for less than a few hours and thus can only cover a small area. More detailed comparisons between the drone and other platforms are listed in Table~\ref{tab:advantages}.

\begin{table}[bt]
    \small
    \centering
    \begin{tabular}{ | c | c | c | c |}
        \hline
               & Drones & Helicopter/Airplane & Satellite \\\hline
        Operating cost & Low    & Moderate     & High \\\hline
        Resolution     & Very high & Very high & High \\\hline
        Low altitude maneuverability & High & Moderate & Poor \\\hline
        Geographic coverage & Localized & Regional & Entire planet \\\hline
        Continuous working time & Short & Moderate & Long \\\hline
        Operation & Easy & Hard & Hard \\\hline
    \end{tabular}
    \caption{Comparison between the drone and its alternatives.}
    \label{tab:advantages}
\end{table}

In recent years, the powerful features of drones have been continuously improved. Thus, drones have been applied in many fields, such as security and surveillance, agriculture and forest, and disaster detection. Furthermore, computer vision methods play an essential role in most drone applications. Analyzing video or images captured by drones has become an emerging application attracting significant attention from researchers in various domains of computer vision. 

Computer vision aims at deriving meaningful information from digital images, videos, and other visual inputs. Most existing research in drone vision focuses on object detection and tracking (\cite{du2018unmanned, yu2020unmanned, zhu2018visdrone}), drone-autonomous navigation such as flight control and visual localization, or some challenges related to remote sensing like camera calibration (\cite{xu2016mosaicking}). On the other hand, one of the essential areas in computer vision and image processing is image super-resolution (SR). However, limited drone vision datasets are proposed, and most public drone vision datasets focus only on one or two specific tasks, such as object detection (\cite{du2018unmanned, yu2020unmanned, zhu2018visdrone}). To the best of our knowledge, none has offered a drone vision dataset for the image SR task. Consequently, developing a drone vision benchmark for the image SR is of great importance to boosting related research.

The target of single image super-resolution (SISR) is to recover the high-resolution (HR) image from low-resolution (LR) observation. Recently, learning-based methods (\cite{dong2014learning, wang2021unsupervised, bhat2021deep, wang2021learning}) have dominated the research of SISR due to the impressive feature extraction capability of deep neural networks, especially convolutional neural networks (CNN). However, most existing methods are trained and evaluated on synthetic datasets. The LR images are generated by applying a simple and uniform degradation (i.e., MATLAB bicubic downsampling) to HR images. Recently, some researchers (\cite{cai2019toward, wang2021unsupervised}) have found that those methods suffer a severe performance drop when applied in a real-world scenario, as the degradation model is more complex in the real-world. It is thus highly desired that we can have a training dataset consisting of real-world, instead of synthetic, LR and HR image pairs (\cite{cai2019toward}). Therefore, some real-world SR datasets have been proposed (\cite{cai2019toward, cai2019ntire, zhang2019zoom, bhat2021deep}). However, all of them are captured on the ground using digital single-lens reflex (DSLR) cameras or mobile phones. Still, the image degradation model is different among different devices and scenarios, which means these datasets can not apply to drone vision.

\section{Objectives}

In this project, we aim to construct a general and real-world dataset of images captured by the drone's bird-view for the image SR task. To the best of our knowledge, no one has provided a drone vision dataset for image SR. Constructing such a drone vision super-resolution (DroneSR) dataset is non-trivial because perfectly aligned LR and HR image pairs are tough to acquire. Therefore, we propose a more flexible procedure for data acquisition. More specifically, we use a drone equipped with two cameras with different focal lengths. We capture images of the same scene using two cameras one by one. The image captured by the camera with a larger focal length is regarded as our HR ground-truth, and the image captured by the other camera is our LR image. In this way, HR and LR image pairs on different resolutions can be collected. However, in addition to the change of field of view (FoV), using two cameras can result in many other challenges in image processing, such as a change of optical center and a change of camera settings. To solve this problem, we first develop an FoV matching algorithm to find matched FoV in LR images. Then we apply local alignment on LR and HR image patches to eliminate misalignment. Finally, we propose a color correction algorithm to match the RGB color in LR and HR image pairs.

Since the texture and content of images are captured at different altitudes, we collect LR and HR image pairs at ten different altitudes in each scene. Recently, researchers found that using RAW data as SR model input can improve the performance of the SR model. Another way to enhance SR performance is to use burst sequence images as input since different burst frames may contain subpixel information (\cite{bhat2021deep, lecouat2021lucas, dudhane2021burst}). Researchers in the SR area are also interested in building an SR model for the arbitrary scale factor. Therefore, we also collect RAW data for HR and LR images and enable burst mode when capturing LR images to benefit others who want to use our dataset for further research.

Our DroneSR dataset contains HR and LR image pairs at different altitudes. We then evaluate existing SR models on our dataset and find that those methods' performance varies among altitudes. This is caused by the texture and content differences at various altitudes. Specifically, we found that images captured at lower altitudes contain more high-frequency details (see Figure \ref{fig: psd}). Since the altitude information is available in our dataset, we propose a new method to feed the altitude information into SR models to build a robust SR model for drone vision at varying altitudes. Finally, we try to apply a few-shot learning setting to our DroneSR dataset and image SR model, so the SR model trained on some altitudes can be quickly adapted to new altitudes.

The object of this project is to propose the first drone vision dataset for image SR and make the image super-resolution methods robust for drone vision at varying altitudes; therefore, the drone can fly higher and still work. To achieve the target, some tasks are identified: 

\begin{itemize}
    \item To build a drone vision dataset consisting of LR-HR images at different altitudes;
    \item To benchmark state-of-the-art (SOTA) SR models based on our dataset and evaluate the domain gap between standard datasets and our dataset.
    \item To propose methods to make image SR models robust for drone vision at various altitudes.
\end{itemize}

\section{Thesis organization}

This thesis consists of seven chapters in the main body, and it is structured as follows. After this introductory chapter, the second chapter briefly reviews the history of drone vision and image SR. 
The detailed procedures for data acquisition are introduced in the third chapter.
The fourth chapter focuses on the pre-processing steps to generate well-aligned image pairs for SR.
Extensive empirical studies in the fifth chapter show domain gaps exist among images captured from different altitudes.
We propose two methods to build a robust image super-resolution network at different altitudes in the sixth chapter.

In the last chapter, the conclusion of the proposed dataset and methods, as well as some plans and suggestions for further research, are discussed.

\chapter{Literature review}

The last few years have witnessed a burst of drone technology and significant improvement in image SR. This chapter firstly gives a brief overview of drone vision. Then we review several important works for learning-based SISR. Finally, we review some recent advances in the image SR dataset.

\section{Drone vision}
Over the past few years, drones are rapidly growing in popularity due to their low altitude maneuverability, low maintenance cost, and easy operation. As a result, they have become central to the platform of various applications, ranging from military security and academic research to daily life. Computer vision methods play an essential role in those drone applications. Those applications can be grouped into three categories. The first group is related to remote-sensing, such as camera calibration and image matching (\cite{xu2016mosaicking}). The second group related to drones' automating flight and stabilization, including pose estimation and obstacle detection, which is an essential feature of a modern drone. Recently, drone companies have used vision systems for this (\cite{al2018survey}).

Compared with the previous two groups, the applications of images and videos captured by drones attract more attention. Some drone-based image or video datasets have been developed for specific tasks such as traffic and crowd detection (\cite{oh2011large, liu2015fast}) for surveillance. The acquired images by drones may be under different ambient conditions in that dataset. As one of the main areas in computer vision, object detection aims to detect instances of semantic objects of a specific class in images or videos. Recently, more and more drone-based image and video dataset has been proposed for object detection task (\cite{hsieh2017drone, xu2019dac}). The UAVDT (\cite{du2018unmanned}) benchmark is a viral database for the object detection and tracking using drone images. This dataset also contains some real-world challenges, such as various weather conditions, flying altitudes, and drone views.

Large datasets can be used to evaluate conventional approaches and apply new methods, especially for learning-based computer vision methods. However, in recent years, although researchers have developed plenty of datasets for drone-based vision, most of them only focus on object detection and other related tasks. In addition, few works provided publicly available datasets for other computer vision tasks, such as image super-resolution, which requires more effort. 

\section{Single image super-resolution}
Traditionally, SISR methods (\cite{chang2004super, timofte2013anchored}) are based on exemplar or dictionary. These methods generate HR images by leveraging the related patches in a large external database. Therefore, the performance of those methods is restricted by the size and content of the external database. Besides, those methods usually take a lot of time for inference.

With the rapid development of deep learning, learning-based methods have achieved notable improvements over early SR methods and dominated the research of SISR because those methods can extract rich feature representations from images.  As a pioneer work, SRCNN (\cite{dong2014learning}) was firstly proposed using a three-layer architecture to learn LR and HR relationships. The SRCNN needs to upscale the LR image to the target size using bicubic interpolation before feeding it into the network. After that, more CNN-based methods have been proposed. VDSR (\cite{kim2016accurate}) and DRCN (\cite{kim2016deeply}) achieved significant improvement over SRCNN by increasing the depth of the network to 20 and using residual learning and recursive learning to speed up training. Following the idea of residual learning, EDSR (\cite{lim2017enhanced}) was proposed with a very wide architecture, and a very deep one MDSR (\cite{lim2017enhanced}) was also proposed with simplified residual blocks. After that, DRRN (\cite{tai2017image}) was proposed by introducing recursive blocks with shared parameters for stable training. Then a persistent memory network called Memnet (\cite{tai2017memnet}) was developed for the SISR task. Combining residual learning and densely connected block, a dense residual network (RDN) (\cite{zhang2018residual}) increased the depth of the network to more than 100. Then, residual channel attention was introduced into SR by RCAN (\cite{zhang2018image}). SR was also explored on the frequency domain (\cite{zhou2019comparative}). Although these methods achieve impressive performance on SR tasks, the hallucinated details (\cite{el2022bigprior,el2021deep,liang2022image}) are often accompanied by unpleasant artifacts. Generative Adversarial Network (GAN) was used to enhance the visual quality of image SR networks. Inspired by relativistic GAN (\cite{jolicoeur2018relativistic}), ESRGAN (\cite{wang2018esrgan}) was proposed and achieved better visual quality with more realistic image textures. Recently, the Transformer is introduced into image SR by SwinIR (\cite{liang2021swinir}). Combining non-local operation and sparse representation, Non-Local Sparse Attention (NLSA, \cite{Mei_2021_CVPR}) was implemented and achieved better performance on SR.

However, most existing SR networks are trained and evaluated on synthetic datasets such as Div2K (\cite{agustsson2017ntire}), Set14 (\cite{zeyde2010single}) and B100 (\cite{martin2001database}). In those datasets, the HR images are high-quality natural images, and the LR counterparts are generated with predefined downsample kernels. The degradation model in this setting is simpler than the one in real-world scenarios, which leads to a severe performance drop when these methods are evaluated on real-world images (\cite{gu2019blind, wang2021unsupervised, shocher2018zero}). To solve such problem, some real-world SR dataset are collected such as RealSR (\cite{cai2019toward, cai2019ntire}), SR-RAW (\cite{zhang2019zoom}), W2S (\cite{zhou2020w2s}) and BurstSR (\cite{bhat2021deep}). BSRNet and BSRGAN (\cite{zhang2021designing}) were proposed leveraging a more complex but practical degradation model considering randomly shuffled blur, downsampling, and noise degradation. Combining an unsupervised degradation representation learning scheme and degradation aware block, DASR (\cite{wang2021unsupervised}) achieved better performance on blind SR task.

Zero-shot learning was introduced in SR to adapt networks to the new degradation model quickly. Researchers have found that adaptive networks achieved better performance, especially for unseen images (\cite{park2020fast}). ZSSR (\cite{shocher2018zero}) was proposed by training a small network at test time from scratch. The training samples of ZSSR were extracted and generated only from the given input image. Inspired by model agnostic meta-learning (MAML, \cite{finn2017model}), MLSR (\cite{park2020fast}) was proposed and effectively handled unknown SR kernels. By finding a generic initial parameter suitable for internal learning, MZSR (\cite{Soh_2020_CVPR}) yields better results with less gradient update.


\section{Recent advances in image super-resolution dataset}

Recently, some researchers found that using RAW data could achieve better performance on real-world SR tasks (\cite{zhang2019zoom}). Therefore, some latest real-world SR datasets provide not only RGB data but also RAW data of LR and HR image pairs (\cite{zhang2019zoom, bhat2021deep}). On the other hand, multi-frame super-resolution (MFSR) (\cite{bhat2021deep, lecouat2021lucas, dudhane2021burst}) usually yield better results than SISR. The input of the MFSR task is a sequence of LR images captured under the burst model. As different burst frames may contain sub-pixel information, this would improve the performance of SR methods. All premonitions methods are designed for fixed scale factors, making SR for arbitrary scale factors become an emerging task. Leveraging meta-learning, Meta-SR (\cite{hu2019meta}) was the first SR method for arbitrary scale factors. Then ArbSR (\cite{wang2021learning}) was proposed by leveraging scale-aware feature adaption blocks and a scale-aware upsampling layer.

\section{Summary}

This chapter briefly reviews the development of drone-based computer vision research and image super-resolution. On the one hand, most existing drone-based vision datasets are designed for object detection and related tasks. To the best of our knowledge, none has proposed a drone vision dataset for the image SR task. On the other hand, the performance of existing SR networks is susceptible to the dataset, specifically, the degradation model between LR and HR images. Therefore, it is significant to propose the first drone vision dataset for image SR.
\chapter{Data acquisition}

This chapter introduces the detailed procedures to acquire our drone vision dataset for super-resolution (DroneSR). We first introduce the platform used in this project in Section \ref{sec: paltform}. Then we explore the effects of different camera settings on captured images and our choice in Section \ref{sec: manual setting} and Section \ref{sec: auto settings}. Finally, we summarize the whole data acquisition procedure in Section \ref{sec: acquisition summary}.

\section{Platform}
\label{sec: paltform}
The platform consists of three parts: the drone itself, the cameras mounted on the drone, and the mobile application to control the drone and cameras.

\subsection{Drone}
\begin{figure}[tb]
    \centering
    \includegraphics[width=.5\linewidth]{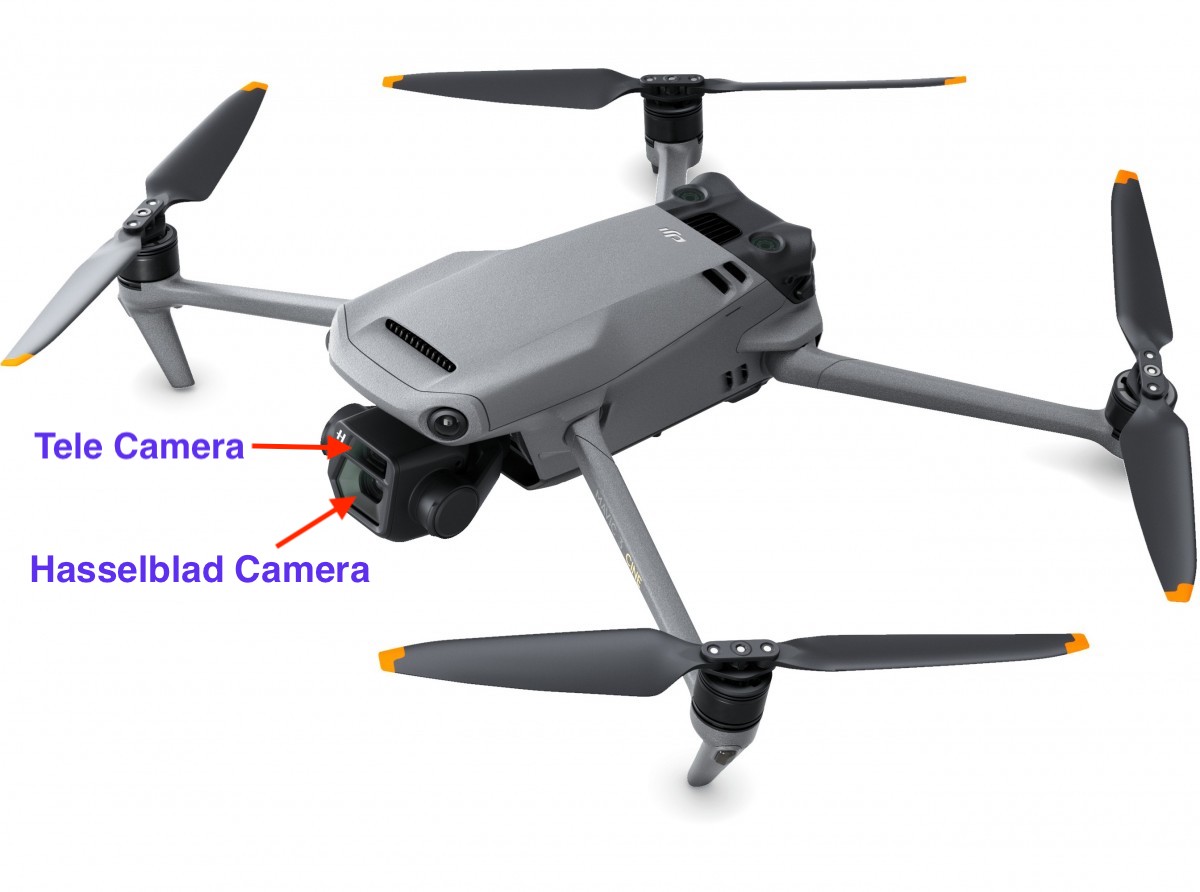}
    \caption{DJI Mavic 3 drone.}
    \label{fig: drone}
\end{figure}

We use DJI Mavic 3 drone (see Figure \ref{fig: drone} for the appearance of the drone) for data acquisition. The drone has 3-axis gimbals to stabilize cameras. The control tile range of cameras is $-90^{\circ}$ to $+35^{\circ}$. In this project, we only focus on the bird-view, which means the camera shoots on the top of objects, so the tile angle is set to be $-90^{\circ}$ throughout the data acquisition process.

The drone utilizes the GNSS (GPS+Galileo+BeiDou), a vision system, and an infrared sensing system to locate and stabilize itself. When the GNSS signal is strong, the drone uses GNSS, and this mode's vertical hovering accuracy range is $\pm0.5m$. On the other hand, if the GNSS signal is weak but the environmental conditions are sufficient, the drone switches to the vision system, and the accuracy range under this situation is $\pm0.1m$.

\subsection{Cameras}

DJI Mavic 3 is equipped with two cameras: one primary camera is a 4/3-in CMOS sensor Hasselblad L2D-20c camera (referred to hereinafter as "Hasselblad camera"); the other is a 1/2-in CMOS sensor Tele camera (referred to hereinafter as "Tele camera"). Some specifications of the two cameras are listed in Table~\ref{tab: cameras}.
\begin{table}[bt]
    \centering
    \begin{tabular}{ | c | c | c | }
        \hline
         & Hasselblad camera & Tele camera \\\hline
        Sensor & 4/3-in CMOS & 1/2-in CMOS \\\hline
        Lens FOV & $84^{\circ}$ & $15^{\circ}$ \\\hline
        Lens Format Equivalent & $24mm$ & $162mm$ \\\hline
        Lens Aperture & $f/2.8\sim f/11$ & $f/4.4$ \\\hline
        Lens Shooting Range & $1m \sim\infty$ & $3m \sim\infty$ \\\hline
        Electronic Shutter Speed & $1/8000s\sim 8s$ & $1/8000s\sim 2s$  \\\hline
        Max Image Size & $5280\times3956$ & $4000\times3000$ \\\hline
    \end{tabular}
    \caption{Specifications of cameras.}
    \label{tab: cameras}
\end{table}

The Tele camera has a larger focal length, so the image captured by the Tele camera can be considered as an optical zoom version of the one from the Hasselblad camera at the same position. Therefore, we choose the image from the Tele camera as our HR ground-truth and the image from the Hasselblad camera as our LR image to collect one LR and HR image pair for our dataset. 

\subsection{DJI Fly App}
DJI Fly is a mobile application provided by the DJI company. Through this application, we can connect our mobile phones with the drone and display the real-time view captured by the camera. In addition, we can set camera parameters in DJI Fly. The available options are listed in Table \ref{tab: fly-app}\footnote{The latest drone firmware allows us to use burst mode on the Tele camera and acquire RAW data from the Tele camera. However, those functions are not provided at the beginning of the project.}.
\begin{table}[bt]
    \centering
    \begin{tabular}{ | c | c | c | }
        \hline
        Parameters & Options (Hasselblad camera) & Options (Tele camera) \\
        \hline
        ISO  & 100, 200, 400, 800, 1600, 3200, 6400 & 100 - 6400 (auto)\\
        \hline
        Shutter & \makecell[c]{1/8000, 1/6400, 1/5000, 1/4000, 1/3200,\\
                                1/2500, 1/2000, 1/1600, 1/1250, 1/1000,\\
                                1/800, 1/640, 1/500, 1/400, 1/320, 1/240,\\
                                1/200, 1/160, 1/120, 1/100, 1/80, 1/60,\\
                                1/50, 1/40, 1/30, 1/25, 1/20, 1/15, 1/12.5,\\
                                1/10, 1/8, 1/6.25, 1/5, 1/4, 1/3, 1/2.5, 1/2,\\
                                1/1.67, 1/1.25, 1", 1.3", 1.6", 2", 2.5",\\
                                3.2", 4", 5", 6", 8"} & 1/8000 - 2 (auto)\\
        \hline
        F-number & \makecell[c]{11.0, 10.0, 9.0, 8.0, 7.1, 6.3, 5.6, \\
                                5.0, 4.5, 4.0, 3.5, 3.2, 2.8} & 4.4 \\
        \hline
        Size & \underline{4:3}, 16:9 & \underline{4:3}\\
        \hline
        Focus Mode & MF(manual focus), \underline{AF}(autofocus) & MF, \underline{AF}\\
        \hline
        Burst Number & 1, 3, 5, \underline{7} & \underline{1}, 3, 5, 7\\
        \hline
        Format & JPG, RAW, \underline{JPG+RAW} & JPG, RAW, \underline{JPG+RAW}\\
        \hline
        Anti-Flicker &  \multicolumn{2}{|c|}{off, \underline{auto}, 50Hz, 60Hz} \\
        \hline
        Peaking Level &  \multicolumn{2}{|c|}{\underline{off}, low, normal, high} \\
        \hline
        White Balance &  \multicolumn{2}{|c|}{\underline{auto}, manual(2000k-10000k)} \\
        \hline
        Explore mode &  \multicolumn{2}{|c|}{off (only use Hasselblad camera), on (above X7, only use Tele camera)} \\
        \hline
        
    \end{tabular}
    \caption{Camera options of DJI Mavic 3 drone in DJI Fly App. The underlined options are chosen for data acquisition.}
    \label{tab: fly-app}
\end{table}

\section{Manual camera settings}
\label{sec: manual setting}
Camera settings are essential for data acquisition. The DJI Fly App provides two modes for us. One is manual mode, and we can change the f-number, ISO, and shutter time of the Hasselblad camera in this mode. The other is auto mode, where we can only choose exposure value (EV). We start with manual mode, test each parameter one by one and try to find the best setting for data acquisition.

\subsection{F-number} 
In an optical system, the f-number ($F$) is defined as the ratio of the system's focal length to the diameter of the entrance pupil:
\begin{equation}
    F = \frac{f}{D}
\end{equation}
where $f$ is the focal length, and $D$ is the diameter of the entrance pupil (effective aperture). 

The depth of field (DoF) defines the focused area within the distance between blurred near-field and far-field objects. The approximate DoF can be given by: 
\begin{equation}
    DoF \approx \frac{2h^2Fc}{f^2}
\end{equation}
where $c$ is the circle of confusion, and $h$ is the distance to the subject (i.e., relative altitude between the drone and ground). 
Object outside DoF suffers from blur. During the data acquisition process, we obtain images of different resolutions using two cameras of different focal lengths. To build a perfect LP and HR image pair, it would be ideal if we made DoF identical, but it is not practical (\cite{zhang2019zoom}). Besides, our Tele camera has a fixed f-number (4.4, see Table~\ref{tab: cameras}), and we can not change the settings of the Tele camera.

\begin{figure}[bt]
    \centering
    \includegraphics[width=\linewidth]{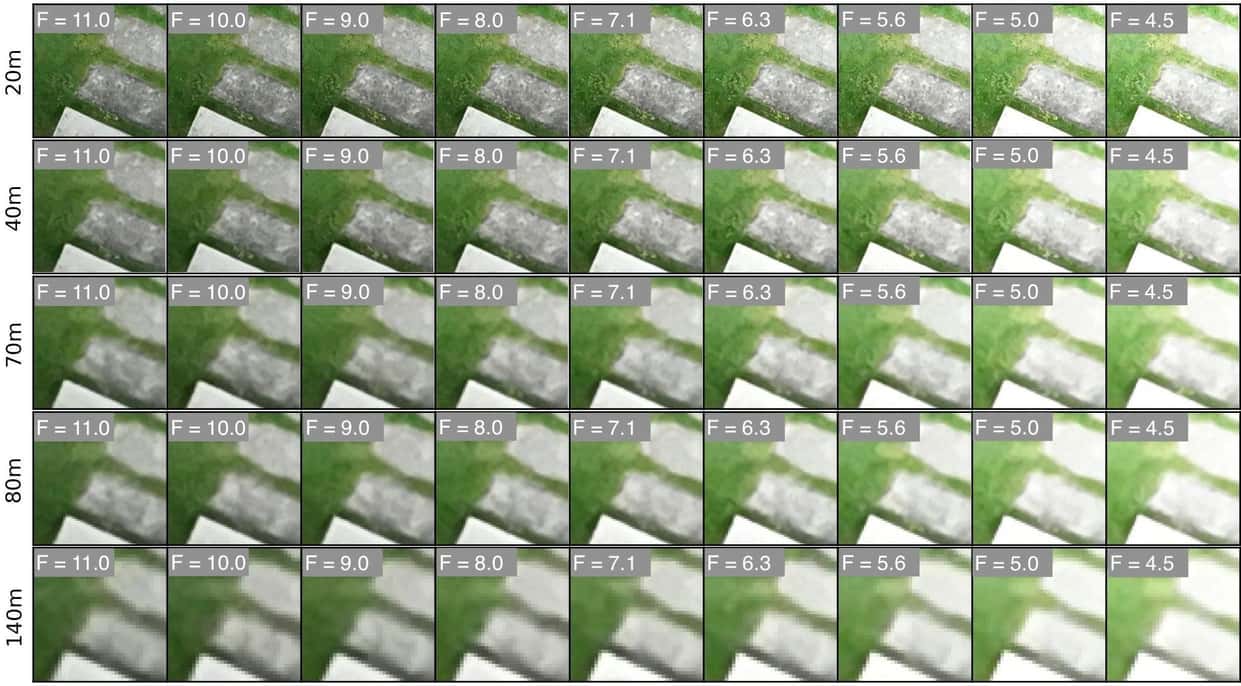}
    \caption{Example patches captured at different relative altitude and different f-number. All of the images were captured with 1/400s shutter time and 1600 ISO. The text on the left shows the relative altitude.}
    \label{fig: aperture}
\end{figure}

To eliminate the influence caused by the DoF mismatch, SR-RAW (\cite{zhang2019zoom}) chose a small aperture size (i.e., large f-number) for all images in the dataset. Since the DoF is related to relative altitude and f-number, we collect a small dataset to show the effect of f-number and relative altitude on captured image (see Figure~\ref{fig: aperture}). In addition, we crop small patches from the original images to show the effects more clearly. As we can see from Figure~\ref{fig: aperture}, the images suffer from more blur as the f-number decreases. However, since the image resolution decrease as relative altitude increase, it is hard to observe the same phenomenon on high-altitude images. As the f-number decreases, we observe the brightness of captured image increase, and it tends to overexposure because aperture size increases and more light comes into the lens. 


\subsection{ISO} 
\begin{figure}[bt]
    \centering
    \includegraphics[width=\linewidth]{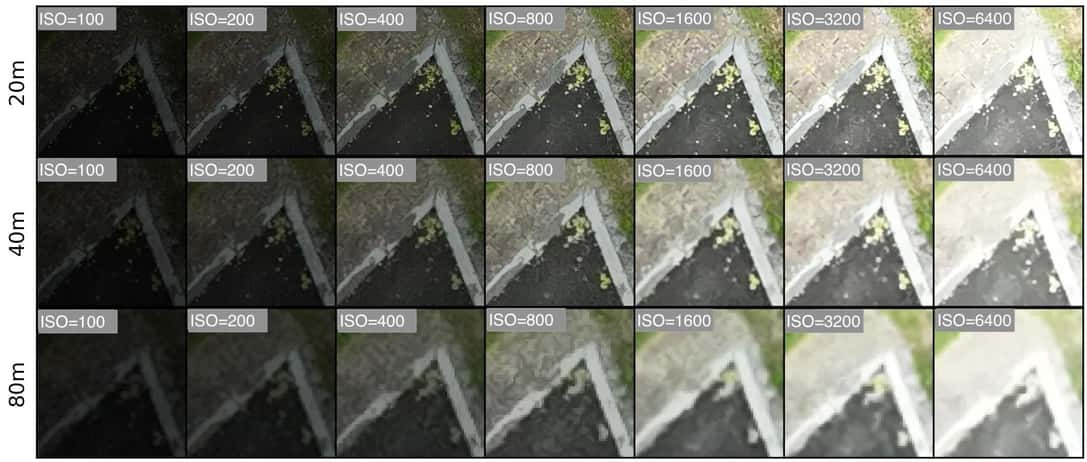}
    \caption{Example patches captured at different relative altitudes and different ISOs. All of the images were captured with 1/500s shutter time and 11.0 f-number. The text on the left shows the relative altitude.}
    \label{fig: iso}
\end{figure}

To achieve proper exposure with a large f-number, we could adjust ISO. Increasing ISO could make the captured photos appear brighter. However, one side effect of larger ISO is that it brings more noise. We also collected a small dataset to show the effect of ISO and relative altitude on captured images (see Figure~\ref{fig: iso}). We can find that as the ISO increase, more noise is introduced, and the brighter the images appear.

\subsection{Shutter}

\begin{figure}[bt]
    \centering
    \includegraphics[width=.6\linewidth]{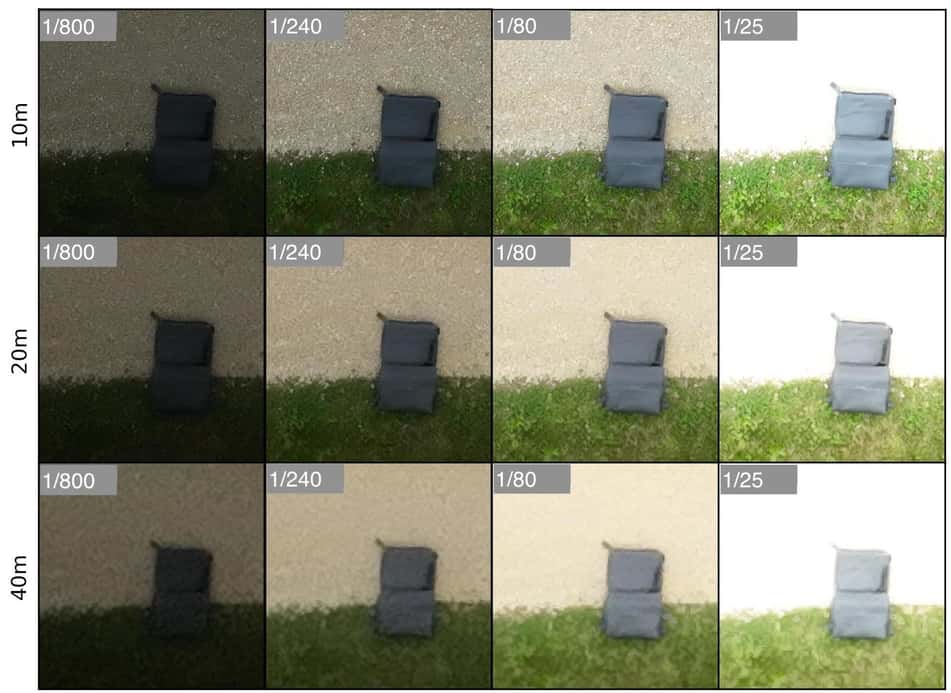}
    \caption{Example patches captured at different relative altitudes and different shutter times. All of the images were captured with 11.0 f-number and 400 ISO. The text on the left shows the relative altitude, and the text in the patches indicates the shutter time (in seconds).}
    \label{fig: shutter}
\end{figure}

Another setting that can be used to adjust the exposure of the captured image is shutter time. The shutter time is defined as the length of time that the camera's sensor is exposed to light when taking a photo. The longer the shutter time, the more light will come into the camera, and the brighter image will appear. However, a longer shutter time leads to a more severe blur due to the drone's vibration. We collected a small dataset to show the effect of shutter time and relative altitude on captured images (see Figure~\ref{fig: shutter}). The results prove that longer shutter time leads to brighter images, thus overexposure and more severe blur.

\subsection{Summary of manual mode}
The exposure is affected by f-number, shutter speed, and ISO. To minimize the DoF difference and blur, we could choose a large f-number. However, a large f-number leads to underexposure. We could increase shutter time and ISO to achieve proper exposure, leading to higher noise or more severe blur. Therefore, it's not practical to manually set all camera parameters at each sense and each altitude.

On the other hand, we need an optical zoom-in image from the Tele camera as the HR ground-truth in our LR and HR image pair. However, we can only choose the auto mode when using the Tele camera. Therefore, we decided on auto mode for both cameras to minimize the brightness difference between LR and HR images.

\section{Auto camera settings}
\label{sec: auto settings}

\begin{figure}[bt]
    \centering
    \includegraphics[width=\linewidth]{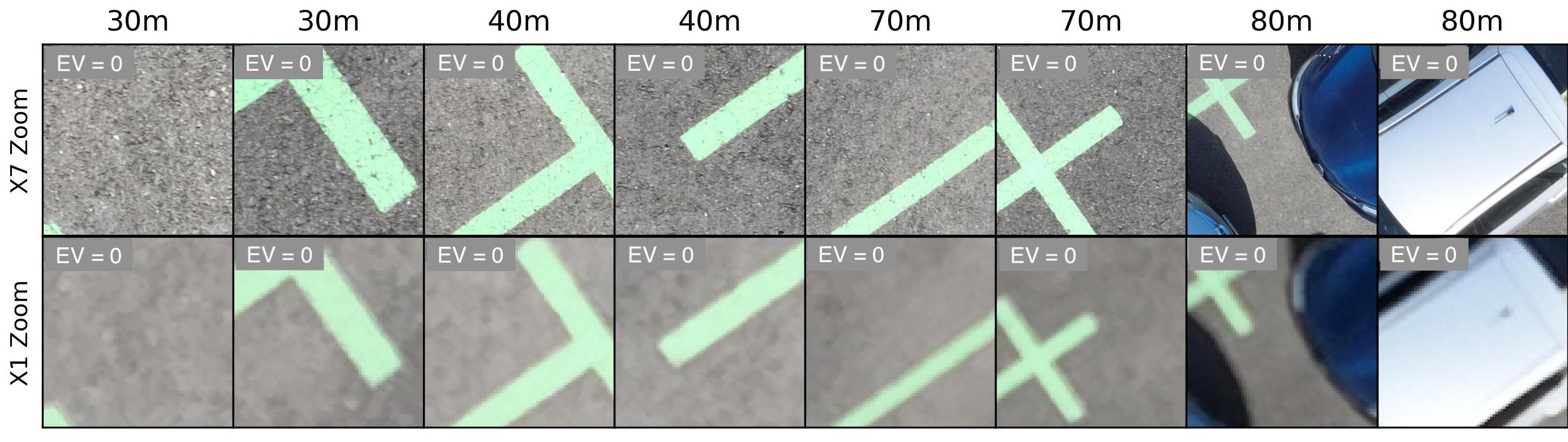}
    \caption{Examples of patches captured by different cameras. The Tele camera takes the top row, and the Hasselblad camera takes the bottom row. The text above each column shows the relative altitude of the drone where the pictures were captured.}
    \label{fig: auto}
\end{figure}

Besides manual mode, the DJI Fly App also provides an auto mode. We can adjust the exposure value (EV) until the image is exposed correctly. In auto mode, when changing EV, the camera will automatically change its shutter time, f-number, and ISO to fit the chosen exposure value. We can use auto mode for both Hasselblad and Tele cameras. We chose the same EV for both cameras. Furthermore, we use the same EV when capturing images for all altitudes. Some examples are shown in Figure~\ref{fig: auto}. 

\section{Compare lower altitude and optical zoom}

\begin{figure}[bt]
    \centering
    \includegraphics[width=\linewidth]{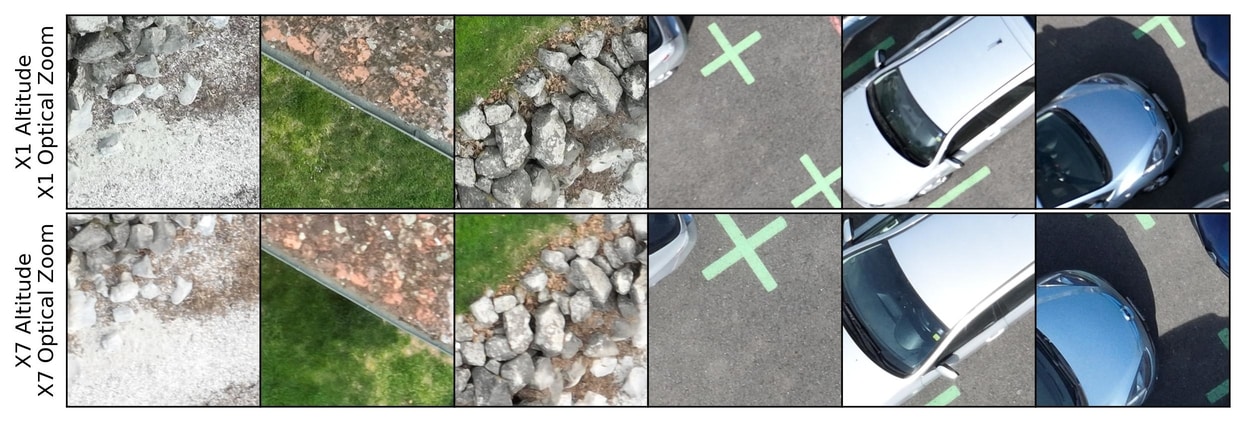}
    \caption{Examples of lower altitude and optical zoom images. Images in the upper row are obtained from lower altitudes (10m or 20m) by the Hasselblad camera; images in the bottom row are obtained from higher altitudes (70m or 140m) by Tele camera (optical zoom).}
    \label{fig: x1-x7}
\end{figure}

Besides using optical zoom to generate LR and HR image pairs, we can collect LR and HR image pairs by taking photos at different altitudes. For example, images taken at lower altitudes can be regarded as HR counterparts of images taken at higher altitudes. We try various strategies to generate the HR and LR image pairs, showing the resulting HR patches in Figure~\ref{fig: x1-x7}. The first row shows patches of HR images captured at lower altitudes, and the second row shows patches of HR images captured using optical zoom. We can obviously find both methods yield good HR images. However, the vertical hovering accuracy of the drone is $\pm0.5m$, which leads to the scale factor varying a lot if we choose a lower altitude image as our HR counterpart. In addition, some challenges, such as perspective distortion, make the further alignment step very complex if we use a lower altitude image. Finally, we choose images captured by Tele camera at the same position as our HR ground-truth. We still collect LR and HR image pairs at different altitudes per scene in case later researchers want to use lower altitude images as HR for the arbitrary scale factor task.

\section{Comparison between burst frames}
\begin{figure}[bt]
    \centering
    \includegraphics[width=\linewidth]{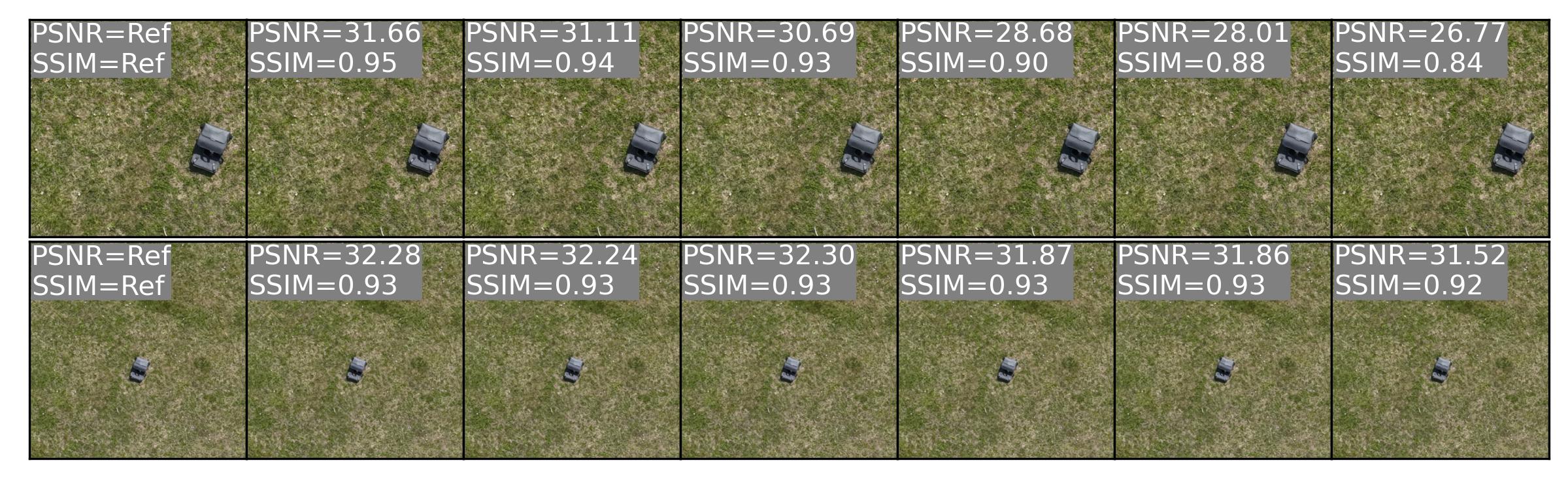}
    \caption{Examples of the burst sequence. The numerical results indicate the differences between burst frames.}
    \label{fig: burst}
\end{figure}

MFSR has been proven to yield better performance than SISR (\cite{bhat2021deep}). We use the burst mode of the Hasselblad camera and take some burst sequence examples. Then we calculate the PSNR and SSIM (\cite{wang2004image}) between each burst frame. The results are shown in Figure~\ref{fig: burst}. Although each burst frame looks very similar, the numerical results show that they are not identical, indicating sub-pixel information may be contained in the sequence. We don't use the burst sequence in this project, but to benefit the later researchers, we choose burst mode when capturing LR images during data acquisition.

\section{Scene selection}
To increase the diversity of our dataset, we need to select as many different scenes as possible and tend to capture images in more flat areas and static scenes to eliminate the mismatch between LR and HR image pairs. For the drone's safety, we also need to collect data where the drone can receive a strong or moderately strong GNSS signal. Due to Swiss drone regulations, we are not allowed to fly the drone close to a group of people.

\section{Summary}
\label{sec: acquisition summary}
In conclusion, we choose the image from the Tele camera as our HR gound-truth and the image from the Hasselblad camera as our LR image to collect one image pair for our dataset, and we chose auto mode for both cameras with the same EV. In this way, we collect image pairs at different altitudes per scene and keep EV the same among all altitudes. The final data acquisition procedures are listed below:

\begin{enumerate}
    \item Take off the drone from some static and flat place with a strong GNSS signal;
    \item Raise the drone to 10m above the starting point;
    \item\label{step: X7zoom} Zoom in X7 (use Tele camera); adjust EV to capture a high-quality image; Save both JPEG and RAW format data;
    \item\label{step: X1zoom} Disable zoom in (use Hasselblad camera); turn on auto mode, keep the same EV as step \ref{step: X7zoom}); turn on burst mode, choose seven frame burst and capture seven frames;
    \item Raise the drone to 20m, 30m, 50m, 40m, 70m, 80m, 100m, 120m, and 140m above the starting point. At each altitude, repeat step \ref{step: X7zoom}) and step \ref{step: X1zoom}), keep EV the same through all altitudes.
\end{enumerate}

Finally, following the setting in BurstSR (\cite{bhat2021deep}), we capture 200 scenes in total, which are split into train, validation, and test sets consisting of 160, 20, and 20 scenes, respectively.








\chapter{Data pre-processing}

This chapter covers the pre-processing steps on the LR and HR image pairs directly captured by the cameras. We first find matched field of view (FOV) in LR images in Section \ref{sec: fov matching}. Then, we apply local alignment on patches of image pairs to eliminate the misalignment problem in Section \ref{sec: local alignment}. Finally, we find the RGB colors of the HR image and LR image are different. To solve this problem, we develop a color correction method in Section \ref{sec: color correction}.

\section{Field of view matching}
\label{sec: fov matching}
Since the images captured by the Hasselblad camera and the Tele camera have different fields of view (FOV), we first crop out the matched FOV from each Hasselblad camera image in the burst sequence. For real-world image FOV matching, \cite{zhang2019zoom, cai2019toward} simply cropped around the center of the LR images because they used only one camera to acquire HR and LR pairs and adjust the focal lengths to achieve optical zoom. They fixed the camera on a tripod, making the center of the HR and LR pair almost not change. However, manually zooming the camera to adjust focal lengths might cause slight camera movement. For alignment, \cite{zhang2019zoom} applied a Euclidean motion model that allows image rotation and translation via enhanced correlation coefficient (\cite{evangelidis2008parametric}). They firstly calculated the alignment between HR and LR RGB images and then apply the alignment to RAW images. \cite{cai2019toward} avoided this problem by using a Bluetooth remote controller. However, those methods do not work well in our dataset as we use two cameras to capture HR and LR images separately, and the vibration of a drone in the air is more severe than human hands and can not be avoided.

\begin{figure}[bt]
    \centering
    \includegraphics[width=.8\linewidth]{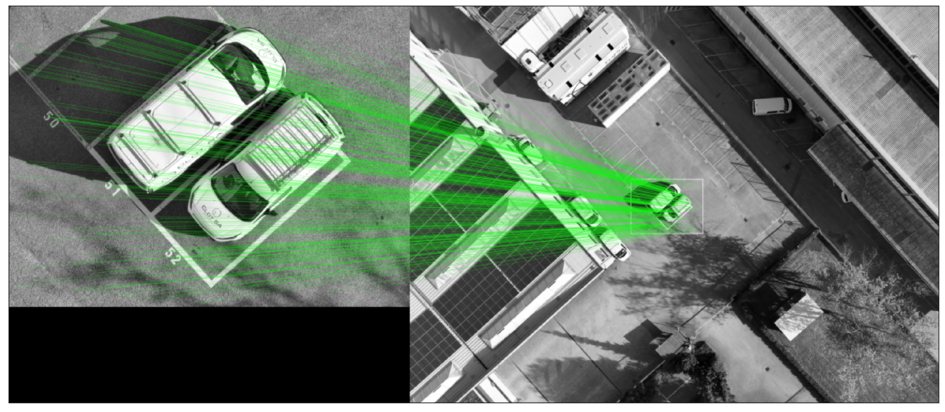}
    \caption{Example of using feature matching (SIFT (\cite{lowe1999object}) and RANSAC (\cite{fischler1981random})) and homography to find FOV in LR. The image on the left is our HR image, and the image on the right is our LR image. The white rectangle in the LR is matched FOV.}
    \label{fig: sift}
\end{figure}

\begin{figure}[bt]
    \centering
    \includegraphics[width=\linewidth]{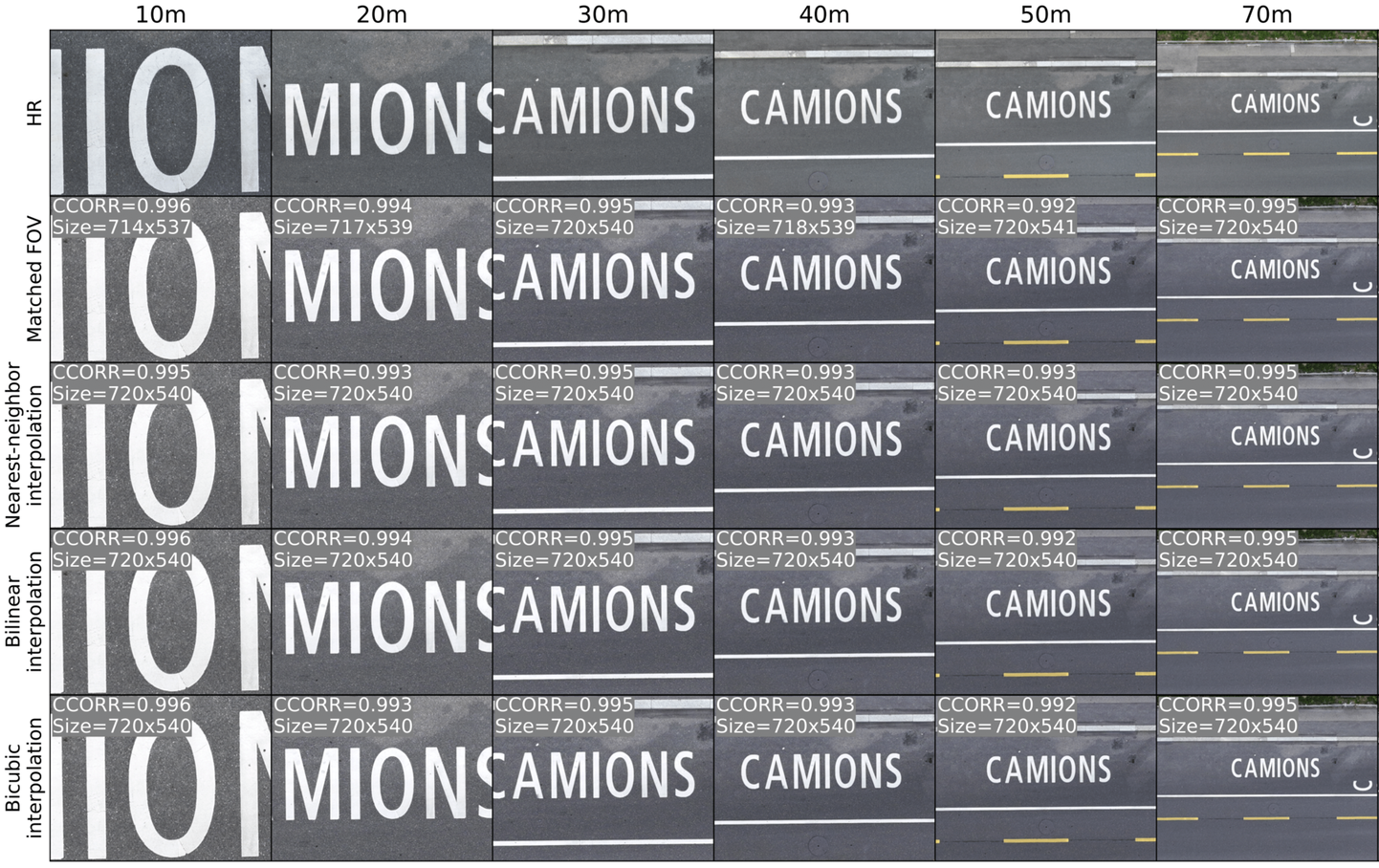}
    \caption{FOV matching examples: the HR image is the full image captured by the Tele camera of size ($4000\times3000s$). The second row shows the matched FOV directly obtained from feature matching and homography. The rest row shows the resized FOV after different interpolation methods. The size of matched FOV is shown in the top-left corner of the image. Finally, the normalized cross-correlation is calculated between matched FOV and downsampled HR images.}
    \label{fig: fov-match}
\end{figure}

\cite{bhat2021deep} also used two cameras to capture HR and LR images separately. One phone camera was used to capture LR, and one digital single-lens reflex (DSLR) camera was used to capture HR. The DSLR camera was also fixed on a tripod, and they held the phone camera by hand just above the DSLR to minimize misalignment. For FOV matching, they estimated the homography between the first image in the LR burst sequence and HR image using SIFT (\cite{lowe1999object}) and RANSAC (\cite{fischler1981random}) (see Figure ~\ref{fig: sift} for an example). We use the same procedure to find matched FOV in our LR images. However, in our case, the drone vibrates not only horizontally but also vertically, which leads to the size of FOV on the LR image also varying (see the second row in Figure \ref{fig: fov-match} for some examples). Although the variation is only a few pixels, this will complicate the rest of the data prepossessing steps and lead to a varying scale factor. We resize the FOV into a fixed size ($720\times540$). Since the size of most matched FOV is around this value. The full size of our HR image is $4000\times3000$. Therefore, the scale factor of our DroneSR dataset is $50/9$. For this resize step, we try different interpolation methods (see the third row to the fifth row in Figure \ref{fig: fov-match}  for more details). We calculate the normalized cross correlation between resized LR images and downscaled HR images and find the results are very similar among all interpolation methods. We think this is because the new size of the LR image is very close to the original one, with only a few pixels difference. So different interpolation gets very similar results. Finally, we choose the nearest-neighbor interpolation method in this step.

\section{Local alignment}
\label{sec: local alignment}
\begin{figure}[tb]
    \centering
    \includegraphics[width=\linewidth]{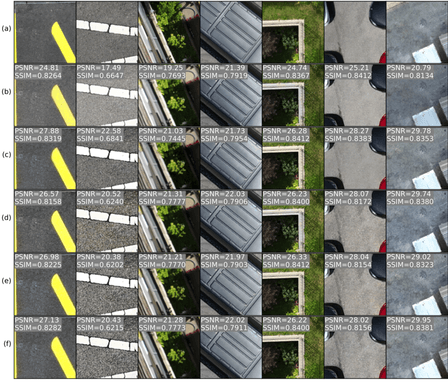}
    \caption{Examples to show the performance of methods for the color mismatch problem. (a): Bicubic downsampled HR images; (b): LR images; (c): Histogram matching; (d): Color transform (\cite{cai2019toward}); (e):  Histogram matching+Color transform; (f): Color transform+Histogram matching. All results are calculated on the Y channel in the transformed YCbCr space.}
    \label{fig: color-correction}
\end{figure}

Like BurstSR (\cite{bhat2021deep}), we extract patches from the LR images in a sliding window manner. To avoid any overlap among patches, we extract $180\times180$ patches from LR images with a stride of 180 pixels. For each patch, we again estimate homography between the LR patch and the corresponding region in the HR image to perform a local alignment. Finally, we filter out HR and LR pairs with a normalized cross-correlation of less than 0.9 (\cite{bhat2021deep}). Some resulted LR patches are shown in Figure \ref{fig: color-correction} row (b), and the corresponding downscaled HR patches are shown in Figure ~\ref{fig: color-correction} row (a).

\section{Color correction}
\label{sec: color correction}
Although we chose the same EV (EV=0) for both Hasselblad and Tele cameras, we still observe the color and luminance difference between HR and LR pairs (see Fig~\ref{fig: color-correction} row (a) and (b) for some examples). 
However, pixel-wise registration is necessary for our dataset. We need to eliminate the mismatched color. To obtain accurate image pair registration, we implemented the pixel-wise color transfer algorithm in RealSR  (\cite{cai2019toward}), which simultaneously considers luminance adjustment. Besides this method, we also try histogram matching \footnote{https://scikit-image.org/docs/dev/auto\_examples/color\_exposure/plot\_histogram\_matching.html\#sphx-glr-auto-examples-color-exposure-plot-histogram-matching-py} between HR and LR pairs, and apply white balance correction (\cite{afifi2019color}) on HR and LR pairs. We find color transfer and histogram matching are more efficient. To compare the performance of different color correction methods, we first calculate the PSNR and SSIM between downsampled HR and LR in a real-world image dataset RealSR (\cite{cai2019toward, cai2019ntire}) (see Table.~\ref{tab: color-matcing-realsr}). Then we try different color correction methods separately and also combine them in a different order. We evaluate those methods on our full dataset, the numerical results are shown in Table~\ref{tab: color-matcing-dronesr} and some samples are shown in Figure \ref{fig: color-correction} row (c)-(f). According to Table~\ref{tab: color-matcing-dronesr}, we choose color transfer and histogram matching in sequential as it achieves the best performance.

\begin{table}[bt]
    \centering
    \begin{tabular}{c|ccc}
        \toprule
        Scale factor & x2           & x3            & x4  \\
        PSNR/SSIM    & 32.25/0.9318 & 30.20/0.9105  & 29.7484/0.9113\\
        \bottomrule
    \end{tabular}
    \caption{PSNR/SSIM between LR images in the RealSR (\cite{cai2019ntire, cai2019toward}) dataset and corresponding downsampled HR images. All results are calculated on the Y channel in the transformed YCbCr space.}
    \label{tab: color-matcing-realsr}
\end{table}
\begin{table}[bt]
    \setlength\tabcolsep{2.5pt}
    \small
    \centering
    \begin{tabular}{c|cccccccccc}
        \toprule
        Altitude (m)                                                &10 & 20 & 30  & 40 & 50 \\
        Origin LR                                                   & 20.12/0.7365 & 20.90/0.7588 & 21.17/0.7733 & 21.44/0.7890 & 21.60/0.7939  \\
        Histogram Matching (HM)                                     & 23.74/0.7598 & 24.59/0.7814 & 25.04/\textbf{0.7959} & 25.35/\textbf{0.8079} & 25.39/0.8109  \\
        Color Transfer (CT)                                         & 23.65/0.7601 & 24.52/0.7813 & 24.91/0.7950 & 25.17/0.8067 & 25.22/0.8102  \\
        HM + CT                                                     & \textbf{23.80}/0.7599 & \textbf{24.63}/0.7809 & \textbf{25.06}/0.7949 & \textbf{25.36}/0.8071 & \textbf{25.42}/0.8103  \\
        CT + HM                                                     & 23.77/\textbf{0.7606} & \textbf{24.63}/\textbf{0.7814} & \textbf{25.06}/0.7954 & \textbf{25.36}/0.8075 & \textbf{25.42}/0.8107  \\
        \hline
        Altitude (m)                                                & 70  & 80 & 100  & 120  & 140 \\
        Origin LR                                                   & 21.52/0.8004 & 21.54/0.8025 & 21.67/0.8114 & 21.99/0.8233 & 22.14/0.8269 \\
        Histogram Matching (HM)                                     & 25.63/\textbf{0.8176} & 25.66/\textbf{0.8192} & 25.87/0.8287 & 26.03/\textbf{0.8392} & 26.07/\textbf{0.8420} \\
        Color Transfer (CT)                                         & 25.47/0.8169 & 25.50/0.8184 & 25.66/0.8276 & 25.79/0.8370 & 25.83/0.8401 \\
        HM + CT                                                     & 25.67/0.8171 & 25.70/0.8185 & 25.92/0.8285 & 26.08/0.8386 & 26.12/0.8417 \\
        CT + HM                                                     & \textbf{25.68}/0.8174 & \textbf{25.71}/0.8186 & \textbf{25.93}/\textbf{0.8290} & \textbf{26.09}/0.8387 & \textbf{26.13}/\textbf{0.8420} \\
        \bottomrule
    \end{tabular}
    \caption{PSNR/SSIM between LR patches with different color correction methods and corresponding downsampled HR patches in our dataset. All results are calculated on the Y channel in the transformed YCbCr space. The best results are in bold.}
    \label{tab: color-matcing-dronesr}
\end{table}

\section{Misalignment analysis}
\label{sec: misalignment analysis}
\begin{figure}[tb]
\centering

\subfloat[Perspective misalignment (L1$\neq$L2).]
{\label{fig: misalignment-perspective}
\includegraphics[width=.5\columnwidth]{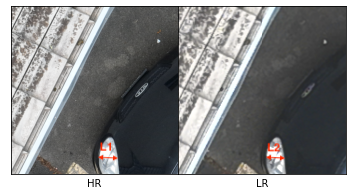}} 
\subfloat[Misalignment caused by movement.]
{\label{fig: misalignment-movement}
\includegraphics[width=.5\columnwidth]{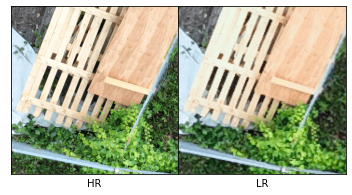}} \\
\subfloat[Resolution alignment ambiguity.]
{ \label{fig: misalignment-ambiguity}
\includegraphics[width=.5\columnwidth]{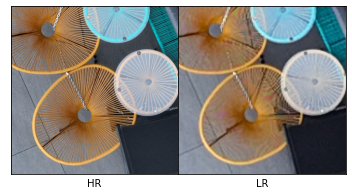}} 
\caption{Examples of noticeable misalignment.}
\end{figure}

Misalignment is unavoidable during data acquisition and is hard to be eliminated by the pre-processing step. By comparing the results in Table \ref{tab: color-matcing-realsr} and Table \ref{tab: color-matcing-dronesr}, our dataset suffers more misalignment than other real-world SR datasets, which is reasonable. Since we capture data with different focal lengths, misalignment is inherently caused by perspective changes. Furthermore, we did not capture the HR and LR images synchronously. There is no absolute flat ground. Thus any vibration of the drone would lead to the movement of the perspective center, which leads to misalignment (Fig. \ref{fig: misalignment-movement}). Besides, it is hard to find static scenes. For example, the wind may cause a slight movement of vegetation (Fig. \ref{fig: misalignment-movement}). Furthermore, sharp details in the HR images cannot be perfectly aligned in the LR image (\cite{zhang2019zoom}, Figure. \ref{fig: misalignment-ambiguity}). The vibration of drones is much more severe than a tripod and human hands, which leads to more severe misalignment problems in our DroneSR dataset. This is inherently caused by the drone and scene and is unavoidable.

\section{Summary}
\begin{figure}
    \centering
    \includegraphics[width=.6\linewidth]{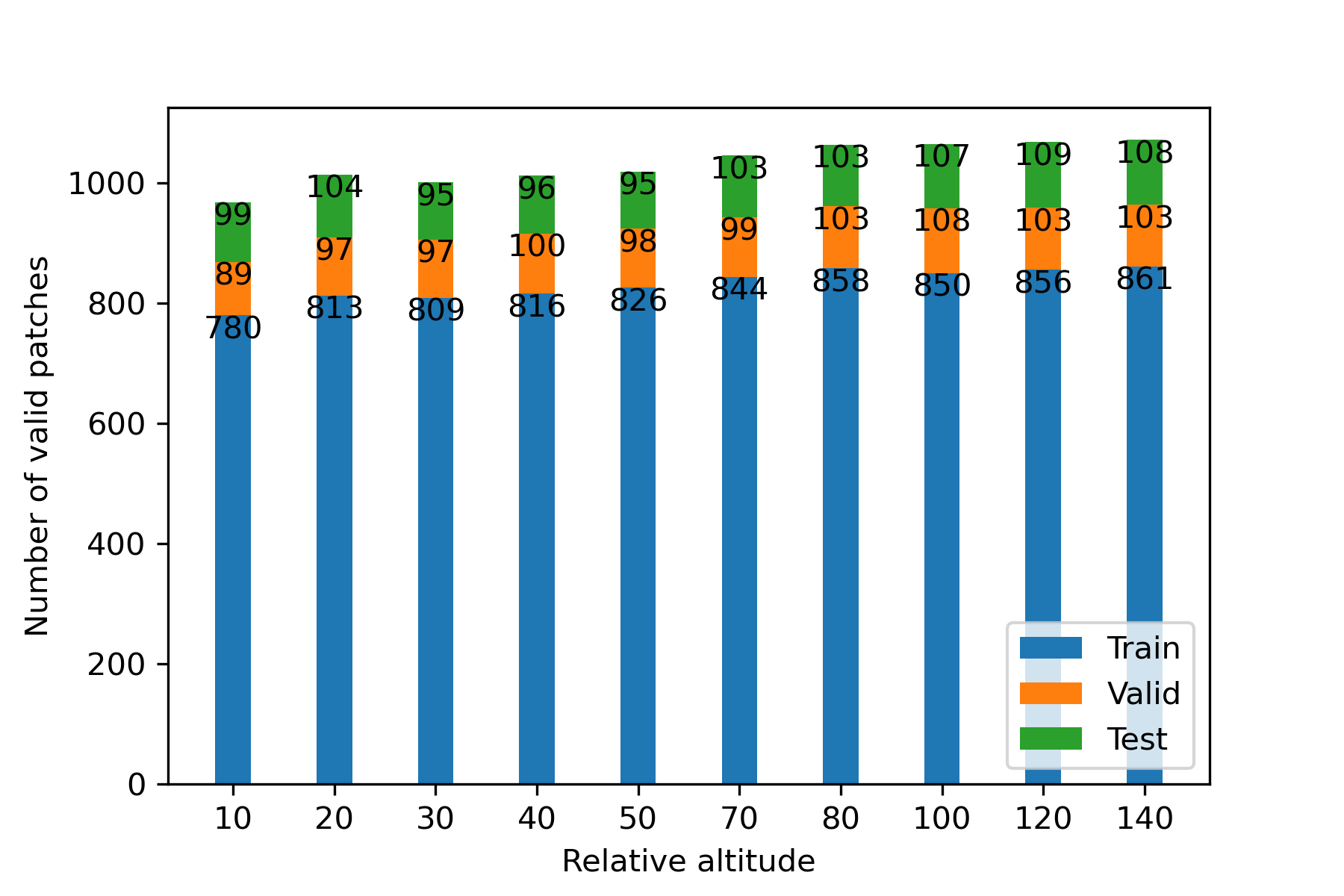}
    \caption{Statistics of our dataset.}
    \label{fig: dataset_statistic}
\end{figure}

The images captured by the Hasselblad and Tele cameras have different FOVs. We first apply FOV matching on the LR image. Then, we use local alignment to eliminate the misalignment between LR and HR pairs and filter out patches with severe misalignment. After that, we notice the color mismatch between HR and LR pairs. Therefore, a color correction algorithm is developed. We visually check some resulted patches and find that misalignment still exists and can not be avoided in a real-world scenario. Due to the inherent property of our platform and scene, our DroneSR dataset suffers more severe misalignment than other real-world SR datasets collected by DSLR with a tripod. Finally, the statistics of the valid number of patches in each split of our dataset are shown in Figure~\ref{fig: dataset_statistic}. Our dataset can be regarded as a balanced dataset among all altitudes.

\chapter{Domain gap observation}

This chapter explores the effects of different altitudes in our DroneSR dataset on SISR networks. We first introduce the degraded performance of pretrained SISR networks in a real-world dataset in Section~\ref{sec: real}. Then we evaluate the pretrained SR networks on our DroneSR datasets at different altitudes. Next, we fine-tune pretrained networks on our dataset. We propose two experimental setups to fine-tune and evaluate those SR networks separately in Section~\ref{sec: fine-tune}. Finally, we summarise and analyze our results in Section~\ref{sec: analysis} for further research.

\section{From synthetic downsampling to real-world degradation}
\label{sec: real}
Most of the existing learning-based SISR networks are trained and evaluated on synthetic dataset such as Div2K (\cite{agustsson2017ntire}), Flickr2K (\cite{lim2017enhanced, timofte2017ntire}), WED (\cite{ma2016waterloo}), FFHQ (\cite{karras2019style}), Set5 (\cite{bevilacqua2012low}), Set14 (\cite{zeyde2010single}), B100 (\cite{martin2001database}) and Urban100 (\cite{huang2015single}), where the LR images are generated by applying some simple and uniform degradation model (i.e., MATLAB bicubic downsampling) to HR images. However, the degradation model of real-world LR images is far more complicated. Many researchers (\cite{cai2019toward}) have found that SISR models trained on synthetic data did not perform well in practical scenarios. In Table \ref{tab: domain gap:real}, we evaluate some SOTA SISR networks (EDSR(\cite{lim2017enhanced}), RDN(\cite{zhang2018residual}), RCAN(\cite{zhang2018image}), ESRGAN(\cite{wang2018esrgan}) and DASR (\cite{wang2021unsupervised})) on some public synthetic datasets (Div2K, Set5, Set14, B100, Urban100) and one real-world dataset called RealSR (\cite{cai2019toward, cai2019ntire}). We downloaded the pretrained SISR networks provided by their authors with a fixed scale factor: $\times4$. All networks were pretrained on Div2K datasets with bicubic downsampling, except that ESRGAN and DASR were pretrained on both Div2K and Flickr2K datasets, and a designed degradation model was used to generate the LR images for DASR. Following the setting in \cite{zhang2018residual}, we calculate the PSNR, SSIM \cite{wang2004image}, on the Y channel of the transformed YCbCr color space. Since the misalignment problem is more severe in our dataset, the pixel-wise PSNR assessment may not work effectively. We introduce LPIPS (\cite{zhang2018unreasonable}) for image quality assessment. Note that LPIPS is used to measure perceptual quality, and a lower LPIPS value means the super-resolved image is more perceptually similar to the ground truth. Besides, we also add gradient-based assessment GMSD (\cite{xue2013gradient}). The higher the GMSD score indicates a larger distortion range, and thus the lower the image perceptual quality.

\begin{table}[bt]
    \centering
    \begin{tabular}{|c|c|c|c|c|c|c|c|}
        \hline
        \multirow{2}{*}{Method} 
        & \multirow{2}{*}{Metric} & \multicolumn{5}{c|}{Public (Bicubic downsampling)} & Real image \\\cline{3-8}
        & & Div2K & Set5 & Set14 & B100 & Urban100 & RealSR \\\hline

        \multirow{4}{*}{Bicubic}
        &PSNR & 28.24 & 28.62 & 26.00 & 26.07 & 23.23 & 27.28 \\\cline{2-8}
        &SSIM & 0.9871 & 0.8264 & 0.8001 & 0.6731 & 0.9084 & 0.9066 \\\cline{2-8}
        &LPIPS & 0.4018 & 0.3376 & 0.4181 & 0.4476 & 0.4037 & 0.4084 \\\cline{2-8}
        &GMSD & 0.0997 & 0.0965 & 0.1175 & 0.1122 & 0.1478 & 0.1614 \\\hline\hline
        
        \multirow{4}{*}{EDSR} 
        &PSNR & 30.74 & 32.37 & 28.50 & 27.75 & 26.63 & 27.64 \\\cline{2-8}
        &SSIM & 0.9951 & 0.9099 & 0.8655 & 0.7428 & 0.9670 & 0.9134 \\\cline{2-8}
        &LPIPS & 0.3238 & 0.2391 & 0.3356 & 0.3743 & 0.2783 & 0.3953 \\\cline{2-8}
        &GMSD & 0.0580 & 0.0414 & 0.0662 & 0.0761 & 0.0759 & 0.1535 \\\hline

        \multirow{4}{*}{RDN}
        &PSNR & 30.62 & 32.24 & 28.42 & 27.68 & 26.34 & 27.64 \\\cline{2-8}
        &SSIM & 0.9950 & 0.9084 & 0.8633 & 0.7398 & 0.9641 & 0.9135 \\\cline{2-8}
        &LPIPS & 0.3296 & 0.2430 & 0.3401 & 0.3788 & 0.2903 & 0.3953 \\\cline{2-8}
        &GMSD & 0.0596 & 0.0418 & 0.0678 & 0.0774 & 0.0802 & 0.1535 \\\hline
        
        \multirow{4}{*}{RCAN}
        &PSNR & 30.73 & 32.52 & 28.57 & 27.77 & 26.72 & 27.65 \\\cline{2-8}
        &SSIM & 0.9950 & 0.9113 & 0.8659 & 0.7438 & 0.9673 & 0.9135 \\\cline{2-8}
        &LPIPS & 0.3235 & 0.2386 & 0.3350 & 0.3741 & 0.2749 & 0.3952 \\\cline{2-8}
        &GMSD & 0.0582 & 0.0416 & 0.0659 & 0.0761 & 0.0756 & 0.1533 \\\hline
        
        \multirow{4}{*}{ESRGAN}
        &PSNR & 28.18 & 30.34 & 26.11 & 25.31 & 24.35 & 27.57 \\\cline{2-8}
        &SSIM & 0.9901 & 0.8672 & 0.7958 & 0.6502 & 0.9458 & 0.9113 \\\cline{2-8}
        &LPIPS & 0.2431 & 0.1953 & 0.2621 & 0.2882 & 0.2274 & 0.3969 \\\cline{2-8}
        &GMSD & 0.0741 & 0.0522 & 0.0834 & 0.0947 & 0.0928 & 0.1551 \\\hline



        
        \multirow{4}{*}{DASR}
        &PSNR & 30.27 & 31.87 & 28.17 & 27.52 & 25.80 & 27.79 \\\cline{2-8}
        &SSIM & 0.9944 & 0.9048 & 0.8599 & 0.7368 & 0.9591 & 0.9178 \\\cline{2-8}
        &LPIPS & 0.3373 & 0.2455 & 0.3479 & 0.3838 & 0.3039 & 0.3880 \\\cline{2-8}
        &GMSD & 0.0635 & 0.0443 & 0.0714 & 0.0791 & 0.0878 & 0.1478 \\\hline

    \end{tabular}
    \caption{Evaluation results of SOTA SR networks on public synthetic and real-world SR datasets. All networks and datasets have a fixed scale factor of 4, We evaluate the results using PSNR, SSIM, LPIPS, and GMSD on the Y channel in the transformed YCbCr space.}
    \label{tab: domain gap:real}
\end{table}

We can obviously find that the performance of the leaning-based SISR model drops on a real-world dataset (ReaSR) if they are trained on synthetic generated HR-LR pairs. Since the degradation model is different among different datasets, we believe there is a domain gap between synthetic SR datasets and real-world SR datasets.

\section{Effects of altitudes on pretrained SISR networks}

In our DroneSR dataset, we capture LR and HR pairs at different altitudes, enabling us to evaluate SISR networks’ performance at different altitudes. In Table \ref{tab: domain gap1} and Table \ref{tab: domain gap2}, we evaluate some SOTA SISR networks (EDSR(\cite{lim2017enhanced}), RDN(\cite{zhang2018residual}), RCAN(\cite{zhang2018image}), ESRGAN(\cite{wang2018esrgan}), SwinIR(\cite{liang2021swinir}), BSRNet(\cite{zhang2021designing}), NLSN (\cite{Mei_2021_CVPR}) and DASR(\cite{wang2021unsupervised})) on public synthetic datasets (Div2K, Set5, Set14, B100, Urban100) and our dataset at different altitudes. The LR images of public synthetic datasets are generated by MATLAB bicubic imresize method based on the scale factor of DroneSR ($\times50/9$). Following the same procedures in previous section, we downloaded the pretrained SISR models the authors gave with a fixed scale factor: $\times4$. With a designed degradation model, the BSRNet was pretrained on Div2K, Flickr2K, WED, and FFHQ datasets. SwinIR and NLSN were pretrained on Div2K datasets with bicubic downsampling. To fit our scale factor, we feed our LR images into SR networks, then upscale the outputs to the target size using bicubic interpolation. We also calculate the PSNR, SSIM (\cite{wang2004image}), LIPIPS (\cite{zhang2018unreasonable}) and GMSD (\cite{xue2013gradient}) on the Y channel of the transformed YCbCr color space. The results are shown in Table \ref{tab: domain gap1} and Table \ref{tab: domain gap2}. 
\begin{table}[bt]
    \small
    \centering
    \begin{tabular}{|c|c|c|c|c|c|c|c|c|c|}
        \hline
        \multirow{2}{*}{Method} & \multirow{2}{*}{Metric} & \multicolumn{5}{c|}{Public} & \multicolumn{3}{c|}{Our} \\\cline{3-10}
        &  & Div2K & Set5 & Set14 & B100 & Urban100 & 10m & 20m & 30m \\\hline

        \multirow{4}{*}{Bicubic}
        &PSNR & 26.56 & 26.37 & 24.30 & 24.62 & 21.82 & 23.12 & 22.69 & 23.39 \\\cline{2-10}
        &SSIM & 0.9593 & 0.7504 & 0.7065 & 0.6013 & 0.7722 & 0.7089 & 0.6836 & 0.6992 \\\cline{2-10}
        &LPIPS & 0.4731 & 0.4149 & 0.4963 & 0.5281 & 0.4851 & 0.5769 & 0.5959 & 0.5836 \\\cline{2-10}
        &GMSD & 0.1435 & 0.1469 & 0.1641 & 0.1565 & 0.1979 & 0.1836 & 0.1983 & 0.1996 \\\hline\hline
        
        \multirow{4}{*}{EDSR} 
        &PSNR & 28.27 & 28.91 & 26.07 & 25.80 & 23.76 & 22.95 & 22.52 & 23.27 \\\cline{2-10}
        &SSIM & 0.9796 & 0.8581 & 0.7870 & 0.6657 & 0.8793 & 0.7099 & 0.6860 & 0.7027 \\\cline{2-10}
        &LPIPS & 0.3911 & 0.2959 & 0.4065 & 0.4498 & 0.3530 & 0.5435 & 0.5605 & 0.5525 \\\cline{2-10}
        &GMSD & 0.1031 & 0.0889 & 0.1181 & 0.1229 & 0.1391 & 0.1880 & 0.2013 & 0.2014 \\\hline
        
        \multirow{4}{*}{RDN}
        &PSNR & 28.20 & 28.73 & 26.06 & 25.75 & 23.62 & 22.96 & 22.53 & 23.28 \\\cline{2-10}
        &SSIM & 0.9791 & 0.8540 & 0.7841 & 0.6626 & 0.8732 & 0.7100 & 0.6861 & 0.7029 \\\cline{2-10}
        &LPIPS & 0.3979 & 0.3010 & 0.4118 & 0.4556 & 0.3679 & 0.5455 & 0.5626 & 0.5540 \\\cline{2-10}
        &GMSD & 0.1050 & 0.0922 & 0.1195 & 0.1246 & 0.1442 & 0.1874 & 0.2008 & 0.2009  \\\hline
        
        \multirow{4}{*}{RCAN}
        &PSNR & 28.27 & 28.78 & 26.11 & 25.81 & 23.82 & 22.93 & 22.51 & 23.26 \\\cline{2-10}
        &SSIM & 0.9795 & 0.8575 & 0.7867 & 0.6668 & 0.8814 & 0.7097 & 0.6858 & 0.7026 \\\cline{2-10}
        &LPIPS & 0.3900 & 0.2971 & 0.4055 & 0.4488 & 0.3472 & 0.5430 & 0.5604 & 0.5520 \\\cline{2-10}
        &GMSD & 0.1031 & 0.0882 & 0.1178 & 0.1229 & 0.1373 & 0.1889 & 0.2021 & 0.2022 \\\hline
        
        \multirow{4}{*}{ESRGAN}
        &PSNR & 26.35 & 26.66 & 24.53 & 24.01 & 22.25 & 22.78 & 22.37 & 23.13 \\\cline{2-10}
        &SSIM & 0.9666 & 0.7976 & 0.7186 & 0.5890 & 0.8372 & 0.7070 & 0.6833 & 0.7000 \\\cline{2-10}
        &LPIPS & 0.3220 & 0.2700 & 0.3445 & 0.3637 & 0.3007 & 0.5316 & 0.5637 & 0.5569 \\\cline{2-10}
        &GMSD & 0.1155 & 0.1093 & 0.1289 & 0.1333 & 0.1476  & 0.1884 & 0.2015 & 0.2014  \\\hline

        \multirow{4}{*}{BSRNet}
        &PSNR & 27.22 & 27.17 & 24.99 & 25.16 & 22.75 & 22.38 & 22.09 & 22.83  \\\cline{2-10}
        &SSIM & 0.9651 & 0.7924 & 0.7309 & 0.6285 & 0.8269 & 0.7021 & 0.6818 & 0.6992  \\\cline{2-10}
        &LPIPS & 0.4573 & 0.3756 & 0.4761 & 0.5175 & 0.4387 & 0.5471 & 0.5584 & 0.5492 \\\cline{2-10}
        &GMSD & 0.1329 & 0.1259 & 0.1568 & 0.1531 & 0.1816 & 0.1935 & 0.2024 & 0.2034  \\\hline
        
        \multirow{4}{*}{SwinIR}
        &PSNR & 28.45 & 28.79 & 26.10 & 25.88 & 24.09 & 22.91 & 22.47 & 23.22  \\\cline{2-10}
        &SSIM & 0.9809 & 0.8630 & 0.7930 & 0.6727 & 0.8946 & 0.7095 & 0.6852 & 0.7019 \\\cline{2-10}
        &LPIPS & 0.3816 & 0.2952 & 0.3996 & 0.4409 & 0.3335 & 0.5451 & 0.5610 & 0.5530 \\\cline{2-10}
        &GMSD & 0.0989 & 0.0879 & 0.1164 & 0.1202 & 0.1291 & 0.1902 & 0.2035 & 0.2034  \\\hline
        
        \multirow{4}{*}{NLSN}
        &PSNR & 28.31 & 28.91 & 26.15 & 25.86 & 24.00 & 22.95 & 22.52 & 23.27  \\\cline{2-10}
        &SSIM & 0.9796 & 0.8607 & 0.7891 & 0.6689 & 0.8862 & 0.7099 & 0.6859 & 0.7027  \\\cline{2-10}
        &LPIPS & 0.3893 & 0.2917 & 0.4037 & 0.4433 & 0.3418 & 0.5440 & 0.5605 & 0.5518 \\\cline{2-10}
        &GMSD & 0.1026 & 0.0868 & 0.1164 & 0.1210 & 0.1330 & 0.1887 & 0.2020 & 0.2021  \\\hline
        
        \multirow{4}{*}{DASR}
        &PSNR & 27.97 & 28.25 & 25.83 & 25.64 & 23.36 & 23.03 & 22.60 & 23.32 \\\cline{2-10}
        &SSIM & 0.9775 & 0.8447 & 0.7769 & 0.6590 & 0.8633 & 0.7106 & 0.6866 & 0.7030  \\\cline{2-10}
        &LPIPS & 0.4067 & 0.3100 & 0.4218 & 0.4614 & 0.3864 & 0.5484 & 0.5658 & 0.5561 \\\cline{2-10}
        &GMSD & 0.1087 & 0.0972 & 0.1232 & 0.1259 & 0.1503 & 0.1871 & 0.2007 & 0.2012  \\\hline
        
    \end{tabular}
    \caption{Evaluation results of SOTA SR networks on public and our dataset at different altitudes. Our dataset has a scale factor of 50/9, and all public datasets are downsampled according to this scale factor using the MATLAB bicubic kernel function. We upscale the networks' output to fit our scale factor for the networks that are not pre-trained on our scale factor. We evaluate the performance using PSNR, SSIM, LPIPS, and GMSD on the Y channel in the transformed YCbCr space.}
    \label{tab: domain gap1}
\end{table}

\begin{table}[bt]
    \small
    \centering
    \begin{tabular}{|c|c|c|c|c|c|c|c|c|}
        \hline
        \multirow{2}{*}{Method} & \multirow{2}{*}{Metric} & \multicolumn{7}{c|}{Our} \\\cline{3-9}
        &  & 40m & 50m & 70m & 80m & 100m & 120m & 140m \\\hline

        \multirow{4}{*}{Bicubic}
        &PSNR & 24.25 & 24.41 & 24.14 & 24.16 & 24.27 & 24.34 & 24.41 \\\cline{2-9}
        &SSIM & 0.7325 & 0.7418 & 0.7456 & 0.7518 & 0.7655 & 0.7668 & 0.7724 \\\cline{2-9}
        &LPIPS & 0.5694 & 0.5727 & 0.5633 & 0.5584 & 0.5572 & 0.5597 & 0.5581 \\\cline{2-9}
        &GMSD  & 0.1903 & 0.1932 & 0.1920 & 0.1909 & 0.1875 & 0.1883 & 0.1872 \\\hline\hline
        
        \multirow{4}{*}{EDSR} 
        &PSNR & 24.07 & 24.29 & 23.98 & 23.99 & 24.11 & 24.17 & 24.24 \\\cline{2-9}
        &SSIM & 0.7346 & 0.7455 & 0.7481 & 0.7542 & 0.7689 & 0.7700 & 0.7758 \\\cline{2-9}
        &LPIPS & 0.5429 & 0.5472 & 0.5388 & 0.5343 & 0.5336 & 0.5364 & 0.5367 \\\cline{2-9}
        &GMSD & 0.1927 & 0.1945 & 0.1947 & 0.1935 & 0.1895 & 0.1902 & 0.1896 \\\hline
        
        \multirow{4}{*}{RDN}
        &PSNR & 24.09 & 24.30 & 24.00 & 24.00 & 24.12 & 24.18 & 24.25 \\\cline{2-9}
        &SSIM & 0.7348 & 0.7456 & 0.7484 & 0.7544 & 0.7690 & 0.7702 & 0.7760 \\\cline{2-9}
        &LPIPS & 0.5438 & 0.5481 & 0.5395 & 0.5351 & 0.5346 & 0.5372 & 0.5371 \\\cline{2-9}
        &GMSD & 0.1924 & 0.1944 & 0.1944 & 0.1933 & 0.1894 & 0.1901 & 0.1893 \\\hline
        
        \multirow{4}{*}{RCAN}
        &PSNR & 24.07 & 24.28 & 23.98 & 23.97 & 24.10 & 24.16 & 24.23 \\\cline{2-9}
        &SSIM & 0.7345 & 0.7454 & 0.7482 & 0.7541 & 0.7687 & 0.7702 & 0.7758 \\\cline{2-9}
        &LPIPS & 0.5423 & 0.5466 & 0.5385 & 0.5347 & 0.5336 & 0.5367 & 0.5368 \\\cline{2-9}
        &GMSD & 0.1934 & 0.1953 & 0.1952 & 0.1945 & 0.1903 & 0.1908 & 0.1902 \\\hline
        
        \multirow{4}{*}{ESRGAN}
        &PSNR & 23.91 & 24.14 & 23.81 & 23.80 & 23.94 & 23.95 & 24.01 \\\cline{2-9}
        &SSIM & 0.7322 & 0.7431 & 0.7453 & 0.7515 & 0.7662 & 0.7667 & 0.7723 \\\cline{2-9}
        &LPIPS & 0.5458 & 0.5520 & 0.5437 & 0.5356 & 0.5346 & 0.5394 & 0.5371 \\\cline{2-9}
        &GMSD & 0.1924 & 0.1944 & 0.1943 & 0.1932 & 0.1894 & 0.1904 & 0.1902 \\\hline

        \multirow{4}{*}{BSRNet}
        &PSNR & 23.61 & 23.87 & 23.60 & 23.62 & 23.70 & 23.77 & 23.86 \\\cline{2-9}
        &SSIM & 0.7293 & 0.7419 & 0.7472 & 0.7537 & 0.7678 & 0.7698 & 0.7762 \\\cline{2-9}
        &LPIPS & 0.5415 & 0.5469 & 0.5406 & 0.5375 & 0.5361 & 0.5393 & 0.5399 \\\cline{2-9}
        &GMSD & 0.1981 & 0.1993 & 0.1988 & 0.1980 & 0.1938 & 0.1945 & 0.1933 \\\hline
        
        \multirow{4}{*}{SwinIR}
        &PSNR & 24.02 & 24.25 & 23.94 & 23.94 & 24.05 & 24.10 & 24.16 \\\cline{2-9}
        &SSIM & 0.7342 & 0.7452 & 0.7479 & 0.7539 & 0.7685 & 0.7694 & 0.7755 \\\cline{2-9}
        &LPIPS & 0.5436 & 0.5470 & 0.5393 & 0.5347 & 0.5342 & 0.5369 & 0.5373 \\\cline{2-9}
        &GMSD & 0.1945 & 0.1961 & 0.1961 & 0.1951 & 0.1909 & 0.1917 & 0.1911 \\\hline
        
        \multirow{4}{*}{NLSN}
        &PSNR & 24.08 & 24.30 & 23.99 & 24.00 & 24.12 & 24.18 & 24.25 \\\cline{2-9}
        &SSIM & 0.7347 & 0.7456 & 0.7484 & 0.7545 & 0.7692 & 0.7703 & 0.7760 \\\cline{2-9}
        &LPIPS & 0.5423 & 0.5459 & 0.5379 & 0.5333 & 0.5328 & 0.5355 & 0.5356 \\\cline{2-9}
        &GMSD & 0.1929 & 0.1950 & 0.1948 & 0.1937 & 0.1896 & 0.1902 & 0.1897 \\\hline
        
        \multirow{4}{*}{DASR}
        &PSNR & 24.14 & 24.35 & 24.05 & 24.07 & 24.18 & 24.25 & 24.31 \\\cline{2-9}
        &SSIM & 0.7350 & 0.7457 & 0.7487 & 0.7547 & 0.7692 & 0.7706 & 0.7763 \\\cline{2-9}
        &LPIPS & 0.5454 & 0.5490 & 0.5407 & 0.5364 & 0.5356 & 0.5382 & 0.5381 \\\cline{2-9}
        &GMSD  & 0.1923 & 0.1944 & 0.1940 & 0.1931 & 0.1892 & 0.1897 & 0.1888 \\\hline
        
    \end{tabular}
    \caption{Evaluation results of SOTA SR networks on public and our dataset at different altitudes. Our dataset has a scale factor of 50/9, and all public datasets are downsampled according to this scale factor using the MATLAB bicubic kernel function. We upscale the networks' output to fit our scale factor for the networks that are not pre-trained on our scale factor. We evaluate the performance using PSNR, SSIM, LPIPS, and GMSD on the Y channel in the transformed YCbCr space.}
    \label{tab: domain gap2}
\end{table}

As we can see from Table \ref{tab: domain gap1} and Table \ref{tab: domain gap2}. The learning-based networks perform worse on our dataset and even worse than bicubic. We believe that this is caused by many factors. Firstly, the quality of the cameras on the drone is not as good as DSLR used in other real-world SR datasets, which leads to a more complicated degradation model. Secondly, the drone vibrates a lot, which leads to a more severe blur and a more complex degradation model. Thirdly, as mentioned in Section \ref{sec: misalignment analysis}, the misalignment in the real-world SR dataset is unavoidable and more severe in our dataset. Finally, the color mismatch problem still exists in our dataset.

On the other hand, the performance of pretrained SOTA SR networks also varies among different altitudes. The content and texture viewed by the drone at different altitudes also vary, making each subset an individual domain, which is different from the train set of SR networks. More detailed analyses are given in Section~\ref{sec: analysis}.

\section{Effects of altitudes on fine-tuned SISR networks}
\label{sec: fine-tune}
\begin{figure}[tb]
    \centering
    \includegraphics[width=.6\linewidth]{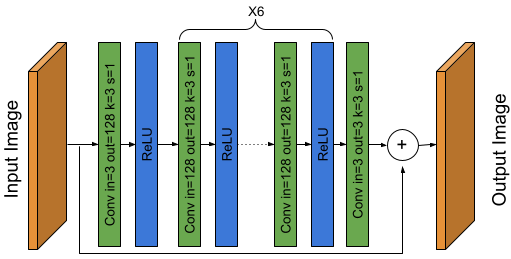}
    \caption{Detailed architecture of our simple CNN network.}
    \label{fig: archi}
\end{figure}

To further evaluate the performance of learning-based networks on our DroneSR dataset at various altitudes and in-depth analyze the domain gap among altitudes. We finetune a simple and fully convolutional network on our dataset. The network contains eight hidden layers, and each layer has 128 channels. We use ReLU activations on each layer. The input LR image is interpolated to the target size before being fed into the network. Following previous CNN-based SR methods (\cite{kim2016accurate, kim2016deeply}), we only learn the residual between the interpolated LR and its HR counterpart. The detailed architecture of our CNN network is shown in Figure \ref{fig: archi}.  We use L1 loss with ADAM (\cite{kingma2014adam}) optimizer. We first pretrain the network on the pubic Div2K dataset with synthetic LR and HR pairs. We randomly crop the HR image into patches of size $300\times300$ with a batch size of 16 for training, and data augmentation is performed through random rotation and flipping. We start with a learning rate of $1\times10^{-4}$, train 1000 epochs in total, and half the learning rate every 200 epochs. 

Then, we propose two setups to fine-tune the pretrained network:

\textbf{Setup 1:} we fine-tune the pretrained network on our DroneSR dataset with data of all altitudes. The initial learning rate is $1\times10^{-5}$. We fine-tune 100 epochs in total and half the learning rate every 20 epochs. The rest hyper-parameters are the same as for pretraining.
\textbf{Setup 2:} we fine-tune the pretrained network on subsets of our DroneSR dataset at each altitude.  The initial learning rate is also $1\times10^{-5}$. Since each subset contains far less data than the whole dataset, we fine-tune 500 epochs in total and half the learning rate every 100 epochs. The rest hyper-parameters are the same as for pretraining.

Finally, we evaluate the performance of those fine-tuned networks as well as the pretrained network at each altitude. Numerical results are shown in Table ~\ref{tab: fine-tune}. we find that fine-tuning model at a specific altitude and then evaluating the model at the same altitude could yield better performance on PSNR. Furthermore, fine-tuning the model on the whole dataset with all altitudes achieves relatively better results than in situations where training altitude and evaluation altitude are largely mismatched, indicating domain gaps exist between subsets of different altitudes. We can also conclude that altitude affects the performance of SR models for drone vision, especially when the model is evaluated at different altitudes from the train set.

\begin{table}[bt]
    \small
    \setlength\tabcolsep{2.5pt}
    \centering
    \begin{tabular}{|c|c|c|c|c|c|c|c|c|c|c|c|}
        \hline
        \multicolumn{2}{|c|}{\multirow{2}{*}{\makecell[c]{PSNR/SSIM/LPIPS}}} & \multicolumn{10}{c|}{Test} \\\cline{3-12}
        \multicolumn{2}{|c|}{}          & 10m & 20m & 30m & 40m & 50m & 70m & 80m & 100m & 120m & 140m \\\hline 
        
        \multicolumn{2}{|c|}{Bicubic}   & \makecell[c]{23.12\\0.7089\\0.5769}
                                        & \makecell[c]{22.69\\0.6836\\0.5959}
                                        &  \makecell[c]{23.39\\0.6992\\0.5836}
                                        &  \makecell[c]{24.25\\0.7325\\0.5694}
                                        &  \makecell[c]{24.41\\0.7418\\0.5727}
                                        &  \makecell[c]{24.14\\0.7456\\0.5633}
                                        &  \makecell[c]{24.16\\0.7518\\0.5584}
                                        &  \makecell[c]{24.27\\0.7655\\0.5572}
                                        &  \makecell[c]{24.34\\0.7668\\0.5597}
                                        &  \makecell[c]{24.41\\0.7724\\0.5581}
                                        \\\hline
        
        \multicolumn{2}{|c|}{Pretrain on Div2K} &  \makecell[c]{22.99\\0.7111\\\textbf{0.5402}}
                                        &  \makecell[c]{22.57\\0.6871\\\textbf{0.5592}}
                                        &  \makecell[c]{23.32\\0.7036\\\textbf{0.5531}}
                                        &  \makecell[c]{24.15\\0.7361\\\textbf{0.5451}}
                                        &  \makecell[c]{24.35\\\textbf{0.7461}\\\textbf{0.5504}}
                                        &  \makecell[c]{24.07\\\textbf{0.7495}\\\textbf{0.5448}}
                                        &  \makecell[c]{24.08\\\textbf{0.7557}\\\textbf{0.5408}}
                                        &  \makecell[c]{24.20\\\textbf{0.7702}\\\textbf{0.5400}}
                                        &  \makecell[c]{24.27\\\textbf{0.7716}\\\textbf{0.5428}}
                                        &  \makecell[c]{24.34\\0.7771\\0.5423}
                                        \\\hline
        
        \multirow{11}{*}{\rotatebox[origin=c]{90}{Fine-tune}}
         
        & All altitudes                 &  \makecell[c]{23.38\\0.7118\\0.5423}
                                        &  \makecell[c]{22.97\\0.6869\\0.5639}
                                        &  \makecell[c]{23.60\\0.6998\\0.5566}
                                        &  \makecell[c]{24.43\\0.7336\\0.5484}
                                        &  \makecell[c]{\textbf{24.58}\\0.7404\\0.5535}
                                        &  \makecell[c]{24.31\\0.7457\\0.5473}
                                        &  \makecell[c]{24.32\\0.7519\\0.5423}
                                        &  \makecell[c]{24.44\\0.7655\\0.5415}
                                        &  \makecell[c]{24.52\\0.7669\\0.5434}
                                        &  \makecell[c]{24.60\\0.7733\\\textbf{0.5421}}
                                        \\\cline{2-12}
                                        
        & 10m                           & \makecell[c]{\textbf{23.44}\\0.7064\\0.5526}
                                        & \makecell[c]{\textbf{23.01}\\0.6805\\0.5767}
                                        & \makecell[c]{\textbf{23.61}\\0.6926\\0.5691}
                                        & \makecell[c]{\textbf{24.45}\\0.7280\\0.5583}
                                        &  \makecell[c]{24.56\\0.7335\\0.5648}
                                        &  \makecell[c]{24.31\\0.7386\\0.5581}
                                        &  \makecell[c]{24.32\\0.7445\\0.5530}
                                        &  \makecell[c]{24.43\\0.7565\\0.5530}
                                        &  \makecell[c]{24.48\\0.7571\\0.5555}
                                        &  \makecell[c]{24.56\\0.7635\\0.5537}
                                        \\\cline{2-12}
                                        
        & 20m                           &  \makecell[c]{23.40\\0.7092\\0.5460}
                                        &  \makecell[c]{22.98\\0.6837\\0.5678}
                                        &  \makecell[c]{\textbf{23.61}\\0.6964\\0.5601}
                                        &  \makecell[c]{24.44\\0.7311\\0.5505}
                                        &  \makecell[c]{24.57\\0.7372\\0.5559}
                                        &  \makecell[c]{24.31\\0.7423\\0.5503}
                                        &  \makecell[c]{24.32\\0.7484\\0.5453}
                                        &  \makecell[c]{24.43\\0.7609\\0.5449}
                                        &  \makecell[c]{24.50\\0.7618\\0.5472}
                                        &  \makecell[c]{24.58\\0.7683\\0.5456}
                                        \\\cline{2-12}
                                        
        & 30m                           &  \makecell[c]{23.36\\0.7097\\0.5428}
                                        &  \makecell[c]{22.95\\0.6844\\0.5643}
                                        &  \makecell[c]{23.59\\0.6975\\0.5571}
                                        &  \makecell[c]{24.41\\0.7315\\0.5483}
                                        &  \makecell[c]{24.56\\0.7382\\0.5532}
                                        &  \makecell[c]{24.28\\0.7435\\0.5473}
                                        &  \makecell[c]{24.29\\0.7494\\0.5422}
                                        &  \makecell[c]{24.41\\0.7627\\0.5416}
                                        &  \makecell[c]{24.48\\0.7637\\0.5439}
                                        &  \makecell[c]{24.56\\0.7702\\0.5424}
                                        \\\cline{2-12}
                                        
        & 40m                           &  \makecell[c]{23.36\\0.7106\\0.5442}
                                        &  \makecell[c]{22.95\\0.6854\\0.5659}
                                        &  \makecell[c]{23.59\\0.6983\\0.5589}
                                        &  \makecell[c]{24.41\\0.7322\\0.5498}
                                        &  \makecell[c]{24.56\\0.7390\\0.5550}
                                        &  \makecell[c]{24.29\\0.7444\\0.5484}
                                        &  \makecell[c]{24.30\\0.7503\\0.5435}
                                        &  \makecell[c]{24.42\\0.7639\\0.5426}
                                        &  \makecell[c]{24.49\\0.7651\\0.5449}
                                        &  \makecell[c]{24.57\\0.7718\\0.5433}
                                        \\\cline{2-12}
                                        
        & 50m                           &  \makecell[c]{23.36\\0.7114\\0.5435}
                                        &  \makecell[c]{22.94\\0.6861\\0.5649}
                                        &  \makecell[c]{23.59\\0.6990\\0.5581}
                                        &  \makecell[c]{24.41\\0.7330\\0.5494}
                                        &  \makecell[c]{24.56\\0.7395\\0.5551}
                                        &  \makecell[c]{24.29\\0.7449\\0.5486}
                                        &  \makecell[c]{24.30\\0.7508\\0.5436}
                                        &  \makecell[c]{24.42\\0.7644\\0.5425}
                                        &  \makecell[c]{24.49\\0.7656\\0.5447}
                                        &  \makecell[c]{24.58\\0.7723\\0.5432}
                                        \\\cline{2-12}
                                        
        & 70m                           &  \makecell[c]{23.34\\0.7133\\0.5422}
                                        &  \makecell[c]{22.93\\0.6884\\0.5625}
                                        &  \makecell[c]{23.58\\0.7012\\0.5553}
                                        &  \makecell[c]{24.41\\0.7347\\0.5479}
                                        &  \makecell[c]{24.57\\0.7414\\0.5536}
                                        &  \makecell[c]{24.30\\0.7470\\0.5469}
                                        &  \makecell[c]{24.31\\0.7529\\0.5421}
                                        &  \makecell[c]{24.43\\0.7667\\0.5408}
                                        &  \makecell[c]{24.50\\0.7679\\0.5431}
                                        &  \makecell[c]{24.58\\0.7743\\0.5414}
                                        \\\cline{2-12}
                                        
        & 80m                           &  \makecell[c]{23.35\\0.7131\\0.5439}
                                        &  \makecell[c]{22.94\\0.6880\\0.5645}
                                        &  \makecell[c]{23.58\\0.7011\\0.5571}
                                        &  \makecell[c]{24.41\\0.7344\\0.5497}
                                        &  \makecell[c]{24.57\\0.7415\\0.5547}
                                        &  \makecell[c]{24.29\\0.7465\\0.5486}
                                        &  \makecell[c]{24.31\\0.7527\\0.5436}
                                        &  \makecell[c]{24.43\\0.7662\\0.5421}
                                        &  \makecell[c]{24.50\\0.7676\\0.5440}
                                        &  \makecell[c]{24.58\\0.7741\\0.5426}
                                        \\\cline{2-12}
                                        
        & 100m                          &  \makecell[c]{23.33\\0.7134\\0.5433}
                                        &  \makecell[c]{22.91\\0.6885\\0.5633}
                                        &  \makecell[c]{23.57\\0.7016\\0.5561}
                                        &  \makecell[c]{24.40\\0.7348\\0.5489}
                                        &  \makecell[c]{24.55\\0.7418\\0.5547}
                                        &  \makecell[c]{24.29\\0.7469\\0.5482}
                                        &  \makecell[c]{24.30\\0.7532\\0.5432}
                                        &  \makecell[c]{24.42\\0.7672\\0.5419}
                                        &  \makecell[c]{24.49\\0.7688\\0.5434}
                                        &  \makecell[c]{24.58\\0.7753\\0.5422}
                                        \\\cline{2-12}
                                        
        & 120m                          &  \makecell[c]{23.33\\0.7130\\0.5436}
                                        &  \makecell[c]{22.91\\0.6882\\0.5637}
                                        &  \makecell[c]{23.57\\0.7018\\0.5563}
                                        &  \makecell[c]{24.40\\0.7347\\0.5496}
                                        &  \makecell[c]{24.55\\0.7419\\0.5552}
                                        &  \makecell[c]{24.29\\0.7472\\0.5487}
                                        &  \makecell[c]{24.30\\0.7535\\0.5442}
                                        &  \makecell[c]{24.42\\0.7674\\0.5428}
                                        &  \makecell[c]{24.50\\0.7690\\0.5446}
                                        &  \makecell[c]{24.58\\0.7753\\0.5434}
                                        \\\cline{2-12}
                                        
        & 140m                          &  \makecell[c]{23.34\\\textbf{0.7155}\\0.5432}
                                        &  \makecell[c]{22.92\\\textbf{0.6907}\\0.5631}
                                        &  \makecell[c]{23.58\\\textbf{0.7038}\\0.5556}
                                        &  \makecell[c]{24.42\\\textbf{0.7368}\\0.5484}
                                        &  \makecell[c]{\textbf{24.58}\\0.7438\\0.5545}
                                        &  \makecell[c]{\textbf{24.32}\\0.7492\\0.5478}
                                        &  \makecell[c]{\textbf{24.33}\\0.7552\\0.5431}
                                        &  \makecell[c]{\textbf{24.45}\\0.7693\\0.5419}
                                        &  \makecell[c]{\textbf{24.53}\\0.7710\\0.5434}
                                        &  \makecell[c]{\textbf{24.61}\\\textbf{0.7772}\\0.5422}
                                        \\\hline
    \end{tabular}
    \caption{Evaluation results of fine-tuned models. We evaluate the performance using PSNR, SSIM, and LPIPS on the Y channel in the transformed YCbCr space. The best results are in bold.}
    \label{tab: fine-tune}
\end{table}

\section{Analysis}
\label{sec: analysis}

\begin{figure}
    \centering
    \includegraphics[width=.8\linewidth]{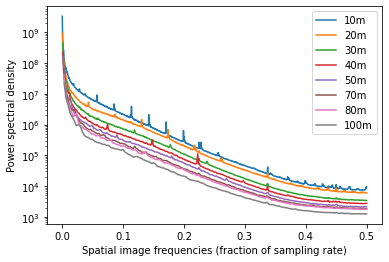}
    \caption{Image frequency content (image PSD) of different altitude.}
    \label{fig: psd}
\end{figure}

As the drone flies higher, the captured image will lose more high-frequency details. To show this, we calculate the power spectrum density (PSD) of our test set images. We crop the center from higher altitudes images to keep the FOV the same among all altitudes. The averaged PSD among all scenes is shown in Figure \ref{fig: psd}.  It is evident that images captured at lower altitudes contain more high-frequency details. This further proves the domain gap between subsets among different altitudes. \cite{el2020stochastic} have found that frequency domain changes would affect SR networks' performance. Therefore, the performance of SR networks varies at different altitudes.

We also observe the mismatch between HR and SR images by calculating the absolute difference between the  bicubic upsampled image and HR image; see Figure \ref{fig: sr-mismatch} for more examples.  As the drone goes up, the mismatch between HR and SR tends to appear in areas with more details.  We also find that bicubic upsampling performs well at edges, which indicates our HR may suffer from severe blur due to the vibration of the drone and explains why bicubic upsampling surpasses pretrained SR networks on our dataset.

In the fine-tuning experiments, we observe that fine-tuning on 10m performs best at lower ($\leq40m$) altitudes in the PSNR, and fine-tuning on 140m performs best at higher altitudes. We illustrate some visual examples of fine-tuning in Section \ref{sec: samples-fine-tune}. We have found the captured HR images at lower altitudes contain more high-frequency information, and these parts affect the final results more at lower altitudes.  Accordingly, the data of 10m includes the most high-frequency details, and thus network fine-tuned on this set tends to perform better on high-frequency information. Therefore, it serves the best at lower altitudes.
On the other hand, the images captured at 140m cover the most expansive area, thus increasing the diversity of the training set. Besides, the HR images suffer less blur from the drone's vibration due to the more considerable distance between the camera and the object. Therefore, the SR network is better trained on a subset of 140m. Furthermore, the network finetuned on the whole dataset is trained on more data with all altitudes. Hence, it performs well among all altitudes. We also illustrate more test samples of fine-tuned models at varying altitudes in Section \ref{sec: samples-fine-tune}.

\section{Summary}

In this chapter, we explore the performance of the existing SR methods on real-world SR datasets. We observe the models trained on synthetic datasets perform less effectively in real-world scenarios. We also find that the performance of those models varies at different altitudes in our DroneSR dataset. Furthermore, we also see the performance of the SR model drops when the model is evaluated at different altitudes from the train set. This motivates us to build a robust SR model for drone vision at varying altitudes.

\chapter{Robust model at varying altitudes}

In this chapter, we propose two methods to build a robust SR model for varying altitudes. The most straightforward and intuitive method is feeding altitude information into the network. We present this method in Section \ref{sec: feed altitude}. To make the SR models quickly to an unknown altitude, we use MAML (\cite{finn2017model}) and propose a one-shot learning scheme for SR models in Section \ref{sec: few-shot}.

\section{Feeding altitude information}
\label{sec: feed altitude}

From the previous analysis, we notice that the altitude information contributes to the performance of SR models. Since we can obtain the altitude information of each LR and HR image pair from the drone, the most intuitive method is to feed the altitude information into our network.

\begin{figure}[tb]
\centering
\subfloat[Overall architecture of our network with altitude information.]
{\label{fig: arch-altitude-overall}
\includegraphics[width=.8\columnwidth]{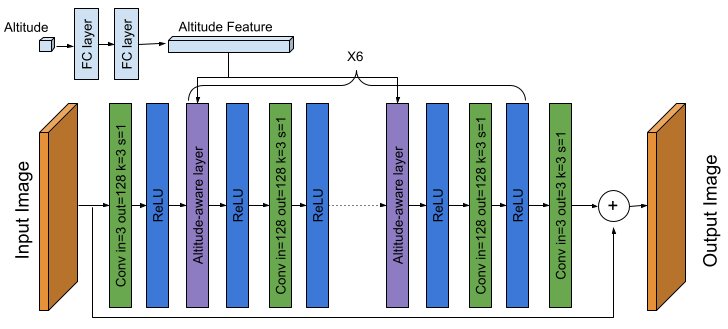}} \\
\subfloat[Altitude-aware layer.]
{\label{fig: arch-aal}
\includegraphics[width=.6\columnwidth]{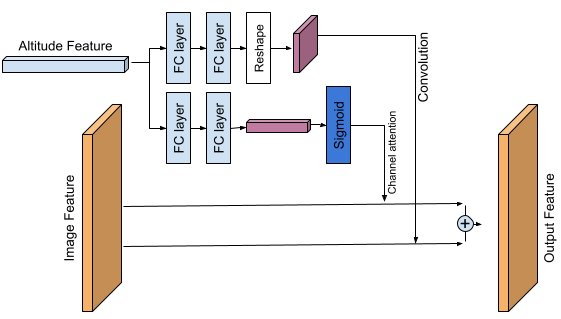}} \\
\caption{Detaild architecture of our fully convolutional network with altitude information.}
\label{fig: arch-altitude}
\end{figure}

Inspired by internal feature manipulation techniques~\cite{el2020blind,lin2021fidelity} and DASR (\cite{wang2021unsupervised}), we modify their degradation-aware convolutional layer to our altitude-aware convolutional layer (AAL).  The AALs are used to adapt the image features conditioned on the altitude. Specifically, the altitude feature is fed into two full-connected (FC) layers and then reshaped to a convolutional kernel $w\in\mathbb{R}^{C\times1\times3\times3}$. Next, the input image feature convolves with the obtained kernel. The resulted image feature is sent to a $1 \times1$ convolutional layer. Meanwhile, the altitude feature is also fed into the other two FC layers, followed by a sigmoid activation to generate weights for channel-wise attention. The input image feature is processed by a channel attention operation using the generated weights. Finally, the results from depth-wise convolution and channel-wise attention are summed together to obtain the output of AAL. The detailed architecture of our AAL is shown in Figure \ref{fig: arch-aal}. We add one AAL between each standard convolutional layer in our original network. Two FC layer is added at the top of the network to encode the altitude information. The detailed architecture of our CNN network with altitude information is shown in Figure \ref{fig: arch-altitude-overall}.

We first pretrain this network under the same procedure as Section \ref{sec: fine-tune}. We still use the Div2K dataset for pretraining. Div2K does not contain altitude information, so we assume all altitudes to be 1. Next, we fine-tune the network on our dataset and feed the altitude into the network. To reduce the variance of altitude, we divided the actual altitude by 80 before sending it. For fine-tuning, we train 100 epochs in total. The initial learning rate is set to be $1\times10^{-5}$ and decreases to half every 20 epochs. We compare the performance of the proposed model with AAL and the original one in Table ~\ref{tab: altitude} and conclude that feeding altitude information into the network significantly improves SR performance.

\begin{table}[bt]
    \small
    \setlength\tabcolsep{2.5pt}
    \centering
    \begin{tabular}{|c|c|c|c|c|c|c|c|c|c|c|}
        \hline
        PSNR/SSIM/LPIPS  & 10m & 20m & 30m & 40m & 50m & 70m & 80m & 100m & 120m & 140m \\\hline 
        
        Bicubic             & \makecell[c]{ 23.12\\0.7089\\0.5769 }
                            & \makecell[c]{ 22.69\\0.6836\\0.5959 }
                            & \makecell[c]{ 23.39\\0.6992\\0.5836 }
                            & \makecell[c]{ 24.25\\0.7325\\0.5694 }
                            & \makecell[c]{ 24.41\\0.7418\\0.5727 }
                            & \makecell[c]{ 24.14\\0.7456\\0.5633 }
                            & \makecell[c]{ 24.16\\0.7518\\0.5584 }
                            & \makecell[c]{ 24.27\\0.7655\\0.5572 }
                            & \makecell[c]{ 24.34\\0.7668\\0.5597 }
                            & \makecell[c]{ 24.41\\0.7724\\0.5581 }
                            \\\hline
        
        Pretrain            & \makecell[c]{ 22.99\\0.7111\\0.5402}
                            & \makecell[c]{ 22.57\\\textbf{0.6871}\\\textbf{0.5592}}
                            & \makecell[c]{ 23.32\\\textbf{0.7036}\\\textbf{0.5531}}
                            & \makecell[c]{ 24.15\\\textbf{0.7361}\\0.5451}
                            & \makecell[c]{ 24.35\\\textbf{0.7461}\\0.5504}
                            & \makecell[c]{ 24.07\\\textbf{0.7495}\\0.5448}
                            & \makecell[c]{ 24.08\\\textbf{0.7557}\\0.5408}
                            & \makecell[c]{ 24.20\\\textbf{0.7702}\\0.5400}
                            & \makecell[c]{ 24.27\\\textbf{0.7716}\\0.5428}
                            & \makecell[c]{ 24.34\\0.7771\\0.5423}
                            \\\hline
        
        All altitudes
                            & \makecell[c]{ 23.38\\\textbf{0.7118}\\0.5423}
                            & \makecell[c]{ 22.97\\0.6869\\0.5639}
                            & \makecell[c]{ 23.60\\0.6998\\0.5566}
                            & \makecell[c]{ 24.43\\0.7336\\0.5484}
                            & \makecell[c]{ 24.58\\0.7404\\0.5535}
                            & \makecell[c]{ 24.31\\0.7457\\0.5473}
                            & \makecell[c]{ 24.32\\0.7519\\0.5423}
                            & \makecell[c]{ 24.44\\0.7655\\0.5415}
                            & \makecell[c]{ 24.52\\0.7669\\0.5434}
                            & \makecell[c]{ 24.60\\0.7733\\0.5421}
                            \\\hline
                            
        \makecell[c]{With\\altitude\\information}
                            & \makecell[c]{\textbf{23.50}\\0.7083\\\textbf{0.5504}}
                            & \makecell[c]{\textbf{23.08}\\0.6850\\0.5654}
                            & \makecell[c]{\textbf{23.70}\\0.7001\\0.5538}
                            & \makecell[c]{\textbf{24.51}\\0.7344\\\textbf{0.5433}}
                            & \makecell[c]{\textbf{24.68}\\0.7422\\\textbf{0.5474}}
                            & \makecell[c]{\textbf{24.38}\\0.7483\\\textbf{0.5402}}
                            & \makecell[c]{\textbf{24.39}\\0.7547\\\textbf{0.5347}}
                            & \makecell[c]{\textbf{24.53}\\0.7700\\\textbf{0.5329}}
                            & \makecell[c]{\textbf{24.59}\\\textbf{0.7716}\\\textbf{0.5360}}
                            & \makecell[c]{\textbf{24.69}\\\textbf{0.7787}\\\textbf{0.5342}}
                            \\\hline
                                        
    \end{tabular}
    \caption{Evaluation results of network with altitude information. "All altitudes" indicates the network mentioned in Section \ref{sec: fine-tune} and fine-tuned on all altitudes. We evaluate the performance using PSNR, SSIM, and LPIPS on the Y channel in the transformed YCbCr space. The best results are in bold.}
    \label{tab: altitude}
\end{table}

\section{Few-shot learning}
\label{sec: few-shot}

Although we can improve drone vision performance by feeding the altitude, the train set must cover all possible evaluation altitudes. However, this is not practical in a real-world application. We cannot collect data from all possible altitudes.  Updating the pretrained model quickly to adapt to a new altitude is essential. However, the naive fine-tune-based update with stochastic gradient descent (SGD) requires a large number of iterations and lots of data from new altitudes.  

To solve this problem, we use a meta-learning technique, MAML (\cite{finn2017model}), to speed up the adaptation procedure. Firstly, we pretrain the SR networks with large external train datasets. We directly use the pretrained model in Section~\ref{sec: fine-tune}. Next, we start meta-learning using MAML, which optimizes the pretrained SR models to enable quick adaptation to the new task. In our dataset, we regard each altitude as an individual task, select some altitudes for meta-training, one new altitude for validation, and evaluate the other altitudes. We use FOMAML (\cite{nichol2018first}) for faster training by ignoring higher-order derivatives. We use one shot, and the details procedure for meta-training is shown in the Algorithm \ref{alg: meta-training}. We conduct 5 gradient updates in the inner loop with a step size of $\alpha=1\times10^{-5}$, and we set $\beta=1\times10^{-4}$.

\begin{algorithm}[bt]
\small
\caption[Meta-training]{Meta-training: one-shot}
\label{alg: meta-training}
\begin{algorithmic}[1]
    \REQUIRE{$f_\theta$: SR model with parameter $\theta$}
    \REQUIRE{$\mathcal{L}$: Loss function}
    \REQUIRE{{$\{D^{1}, D^{2}, ... D^{H}\}$: Subsets of training set at different altitude}}
    \REQUIRE{$\alpha, \beta$: Hyper-parameters}
    \STATE Pretrained model $f_\theta$
    \WHILE{not converge}
        \FOR{each $D^{h}\in\{D^{1}, D^{2}, ... D^{H}\}$}
        \STATE Sample $2$ HR-LR pairs from $D^{h}$: $HR_1^h$, $LR_1^h$, $HR_2^h$, $LR_2^h$
        \STATE Evaluate $\nabla\mathcal{L}(f_\theta(LR_1^h, HR_1^h))$ using $\mathcal{L}$
        \STATE Compute adapted parameters with gradient descent $\theta^h=\theta-\alpha\nabla\mathcal{L}(f_\theta(LR_1^h, HR_1^h))$
        \ENDFOR
        \STATE Update $\theta=\theta-\beta\nabla_\theta\sum_{h=1}^H\mathcal{L}(f_{\theta^h}(LR_2^h), HR_2^h)$
    \ENDWHILE
\end{algorithmic}
\end{algorithm}

For inference, we randomly select one LR and HR pair from the test altitude and update the model with 5 gradient updates like training. The updated model is used for standard inference. We evaluate the performance using different altitudes for meta-training, and the results are shown in Talbe~\ref{tab: meta}. We find that the pretrained SR networks can be quickly updated to the new altitude (140m) when it is meta-trained from 10m to 100m. Although, it does not perform well when meta-trained on 10m to 80m datasets. We think the performance is limited by the size of our dataset and the complexity of our network. Nevertheless, the SR model has the potential to adapt to new altitudes efficiently after meta-training.

\begin{table}[bt]
    \footnotesize
    \setlength\tabcolsep{2.5pt}
    \centering
    \begin{tabular}{|c|c|c|c|c|c|c|c|c|c|c|c|}
        \hline
        Train altitudes(m) & Valid altitudes(m) & 10m & 20m & 30m & 40m & 50m & 70m & 80m & 100m & 120m & 140m \\\hline

        \makecell[c]{10, 20, 30, 40,\\50, 70, 80, 100} & 120 
                            & \textbf{23.52}
                            & \textbf{23.12}
                            & 23.47
                            & 24.45
                            & 24.61
                            & 24.32
                            & 24.27
                            & 24.48
                            & 24.34
                            & 24.51
                            \\\hline
        
        
        10, 20, 30, 40, 50, 70, 80 & 100 
                            & 23.52
                            & 23.11
                            & 23.43
                            & 24.39
                            & 24.55
                            & 24.22
                            & 24.18
                            & 24.36
                            & 24.24
                            & 24.40
                            \\\hline

        \multicolumn{2}{|c|}{Bicubic}
                            & 23.12
                            & 22.69
                            & 23.39
                            & 24.25
                            & 24.41
                            & 24.14
                            & 24.16
                            & 24.27
                            & 24.34
                            & 24.41
                            \\\hline
        
        \multicolumn{2}{|c|}{Pretrain on Div2K}
                            & 22.99
                            & 22.57
                            & 23.32
                            & 24.15
                            & 24.35
                            & 24.07
                            & 24.08
                            & 24.20
                            & 24.27
                            & 24.34
                            \\\hline
        
        \multicolumn{2}{|c|}{All altitudes}
                            & 23.38
                            & 22.97
                            & 23.60
                            & 24.43
                            & 24.58
                            & 24.31
                            & 24.32
                            & 24.44
                            & 24.52
                            & 24.60
                            \\\hline
        
        \multicolumn{2}{|c|}{With altitude information}      
                            & 23.50
                            & 23.08
                            & \textbf{23.70}
                            & \textbf{24.51}
                            & \textbf{24.68}
                            & \textbf{24.38}
                            & \textbf{24.39}
                            & \textbf{24.53}
                            & \textbf{24.59}
                            & \textbf{24.69}
                            \\\hline
                                        
    \end{tabular}
    \caption{Evaluation results of one-shot learning. "All altitudes" indicates the network mentioned in Section \ref{sec: fine-tune} and fine-tuned on all altitudes. We evaluate the performance using PSNR on the Y channel in the transformed YCbCr space. The best results are in bold.}
    \label{tab: meta}
\end{table}

\section{Analysis}

Although feeding the altitude information into the network can improve image SR performance, it increases the complexity of the network, thus increasing the inference time. Besides, the SR network still needs a few steps of gradient update to fit the new altitude during the meta inference phase, which significantly boosts the inference time. We evaluate the time complexities for different methods, and the results are shown in Table \ref{tab: runtime}. We measure time on the environment of the NVIDIA Titan X GPU. Although we only use one shot and five gradient updates for meta-inference, it is still far more time-consuming than normal inference. We also illustrate more test samples of robust models at varying altitudes in Section \ref{sec: samples-robust}.

\begin{table}[bt]
    \small
    \centering
    \begin{tabular}{| c | c |}
        \hline
        Methods & Time (sec) \\\hline
        Simple CNNs & $3.62\times10^{-3}$ \\\hline
        With altitude information  & $3.76\times10^{-3}$ \\\hline
        One-shot learning & 0.31 \\\hline
    \end{tabular}
    \caption{Comparisons of the time complexity of inference for robust methods at varying altitudes of 180×180 LR image.}
    \label{tab: runtime}
\end{table}

\section{Summary}

We proposed two methods to build a robust SR model for varying altitudes. The first feeds altitude information into the network, and the second leverages a one-shot learning scheme, making the SR models quickly adapt to unseen altitudes. Empirical studies show that both methods are efficient. Feeding altitudes into the network improves the performance, and the meta-trained SR network can quickly fit unseen altitudes.
\chapter{Conclusion and future plan}

In this thesis, we propose the first image SR dataset for drone vision. The LR and HR image pairs are captured by two cameras on the drone with different focal lengths. We set the same exposure value for both cameras and collected image pairs at different altitudes.  After that,  we propose the data preprocessing procedures to generate well-aligned HR and LR image pairs for image SR. 

Extensive empirical studies show that pretrained SOTA SR networks suffer a performance drop on our dataset because the degradation model used to train those networks is far simpler than ours. The performance of SR networks varies among altitudes, and we also observe images captured at lower altitudes contain more high-frequency details. Then, we conclude that domain gaps exist among LR and HR pairs from different altitudes.

Finally, we propose two methods to build a robust SR model at varying altitudes. The first feeds altitude information into the network through altitude-aware layers. The second uses one-shot learning to adapt the SR model quickly to unknown altitudes. Experiments show the proposed method can efficiently improve the performance of SR networks at varying altitudes. However, the SR network used in this thesis is very simple, and more complex SR networks remain in our future research.

As an essential technique in computer vision and image processing, image SR can improve other downstream tasks, such as classification and segmentation, which are widely used in many drone applications. Therefore, the benefit of drone base SR on other drone vision tasks can be explored in the future. On the other hand, datasets are critical for learning-based computer vision techniques. Our dataset contains burst LR sequences and RAW data, which benefits further research. We believe our dataset will encourage more researchers to delve into image SR for drone vision.

\appendix
\chapter{More visual examples}

\section{SR matching visualisation}

\begin{figure}
    \centering
    \includegraphics[width=\linewidth]{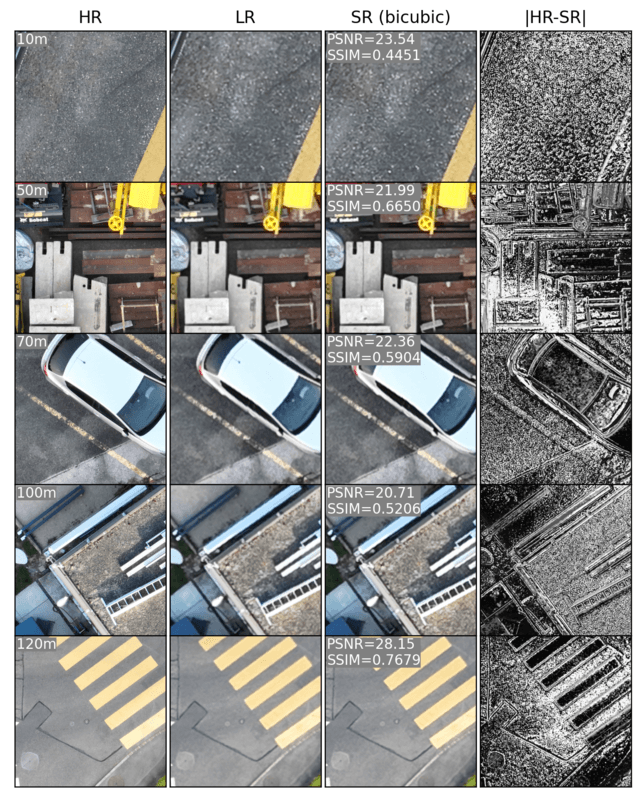}
    \caption{Visualisation of mismatching between HR and SR images.}
    \label{fig: sr-mismatch}
\end{figure}

\section{Test samples of fine-tuning}
\label{sec: samples-fine-tune}
\begin{figure}[tb]
\centering
\subfloat[HR]
{\includegraphics[width=.45\columnwidth]{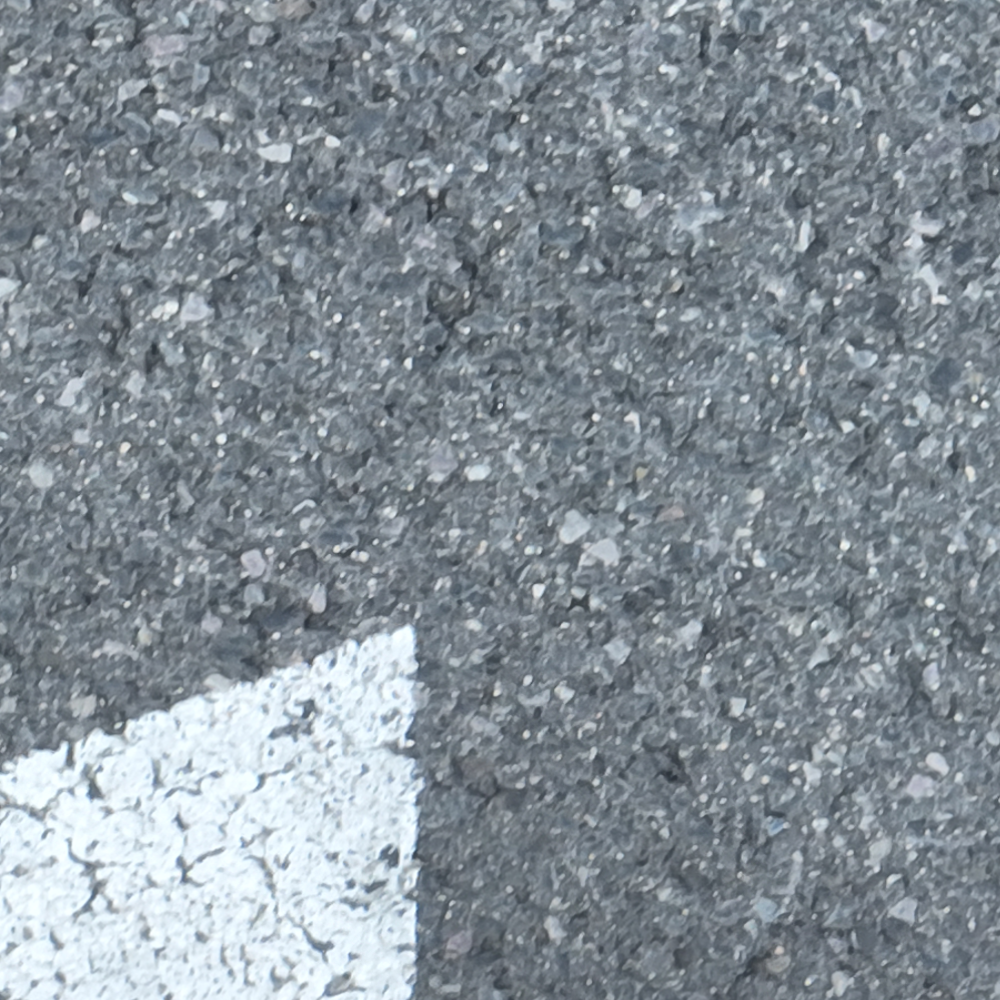}} \hfill
\subfloat[LR]
{\includegraphics[width=.45\columnwidth]{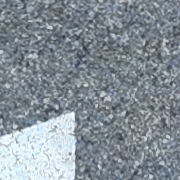}} \\
\subfloat[Pretrain (PSNR: 21.64, SSIM: 0.6709)]
{\includegraphics[width=.45\columnwidth]{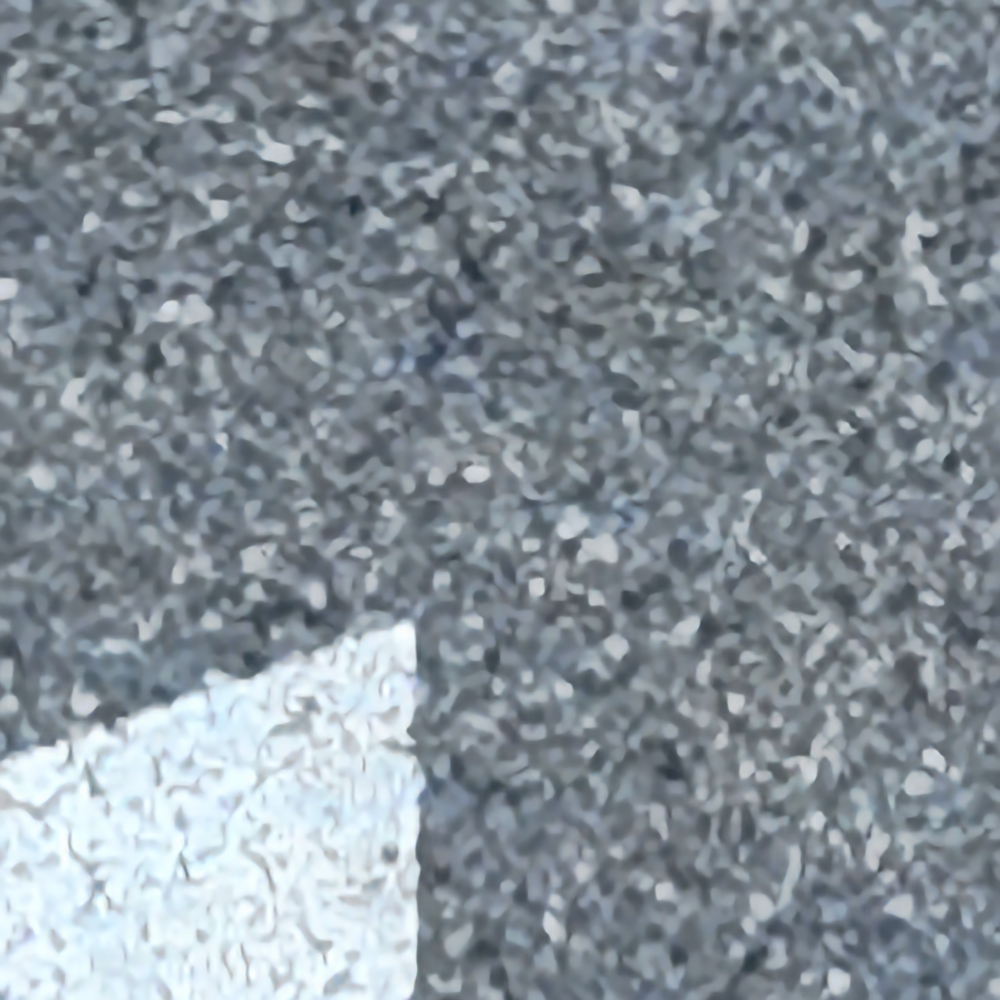}} \hfill
\subfloat[Fine-tune on all altitudes (PSNR: 22.13, SSIM: 0.6517)]
{\includegraphics[width=.45\columnwidth]{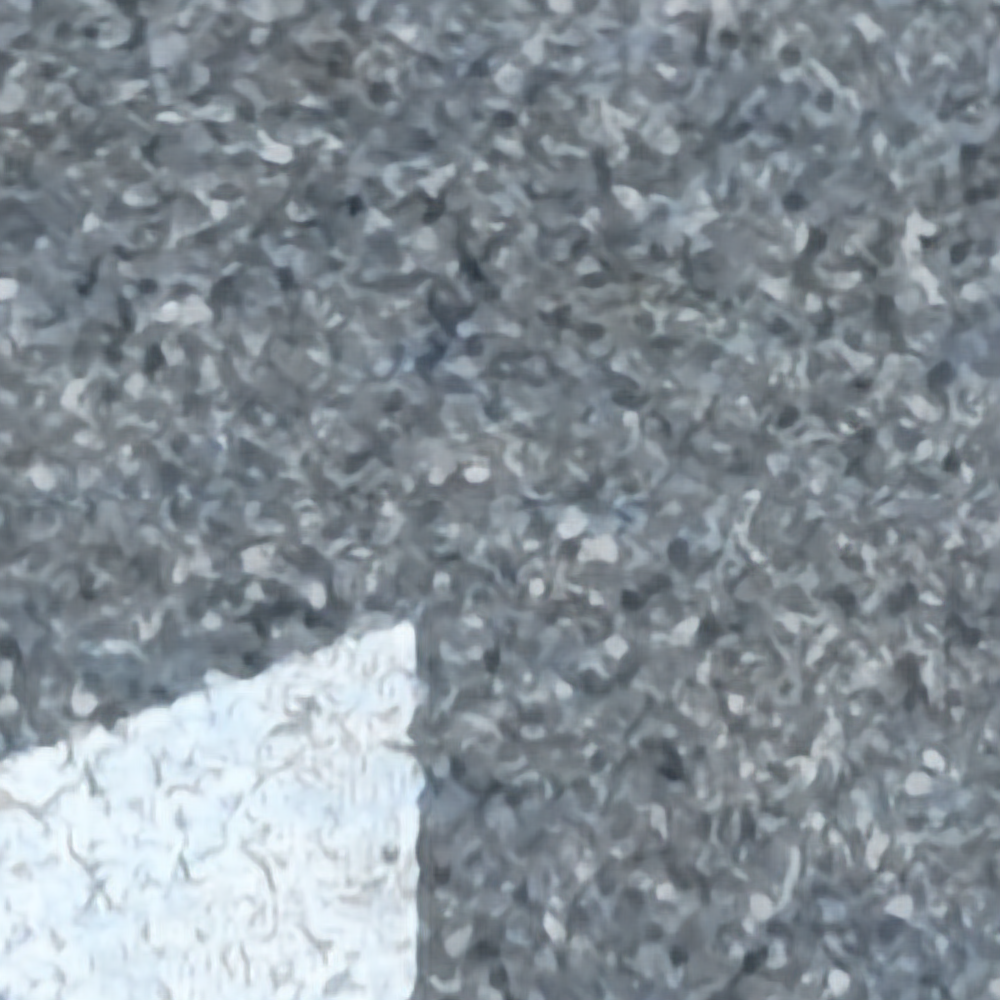}} \\
\subfloat[Fine-tune on 10m (PSNR: 22.15, SSIM: 0.6433)]
{\includegraphics[width=.45\columnwidth]{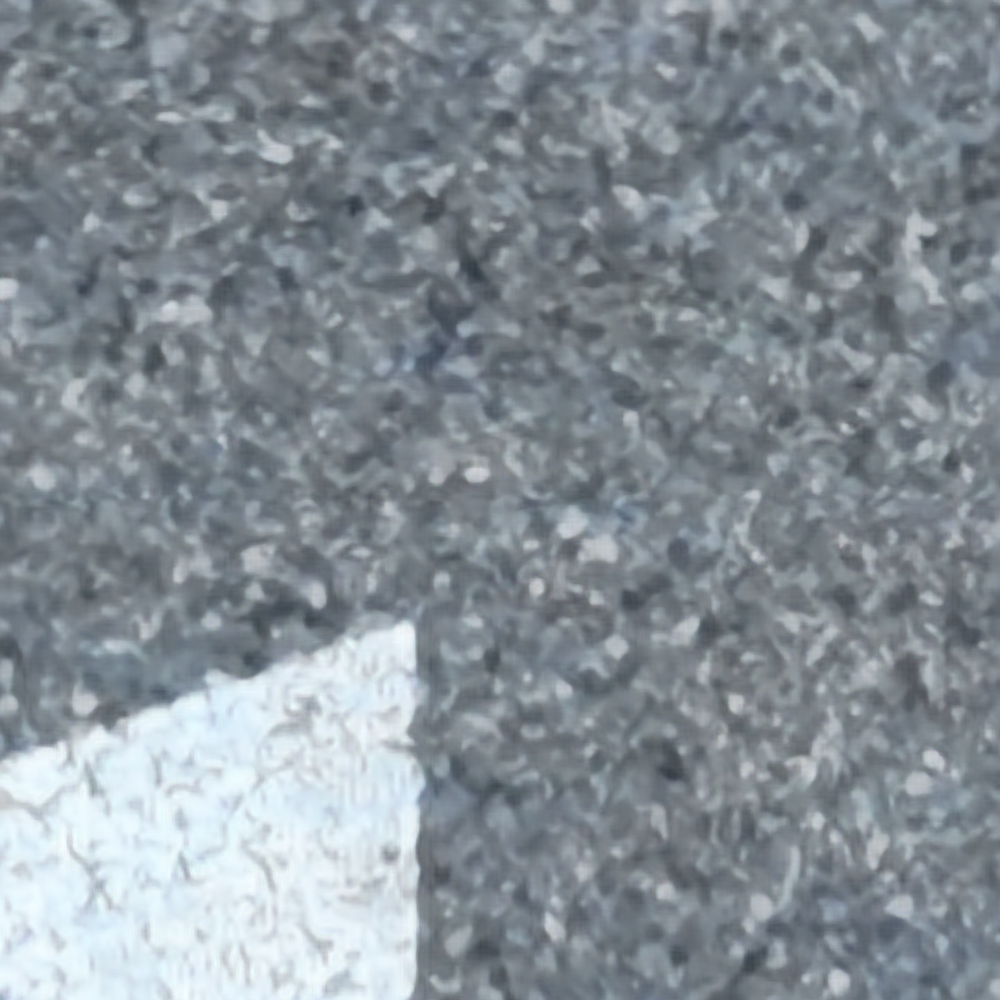}} \hfill
\subfloat[Fine-tune on 140m (PSNR: 22.09, SSIM: 0.6622)]
{\includegraphics[width=.45\columnwidth]{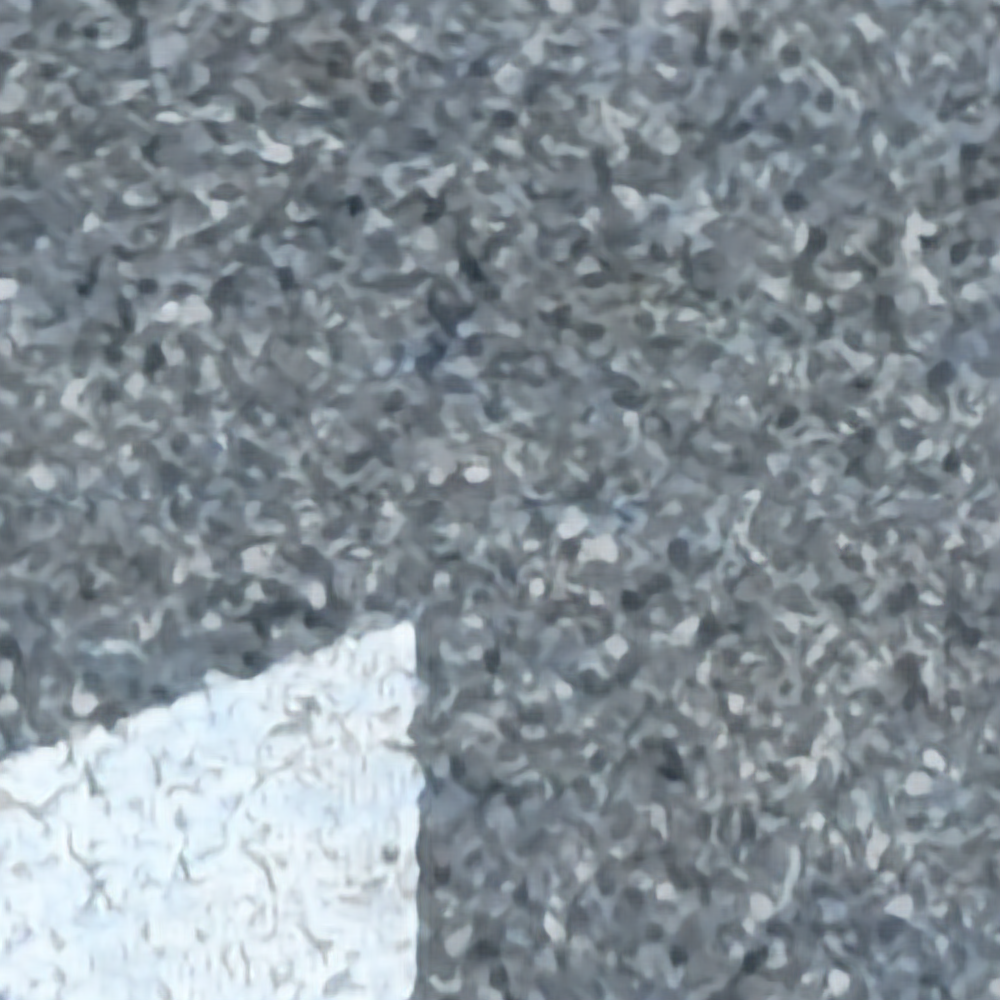}} \\
\caption{Test sample at 10m altitude}
\end{figure}



\begin{figure}[tb]
\centering
\subfloat[HR]
{\includegraphics[width=.45\columnwidth]{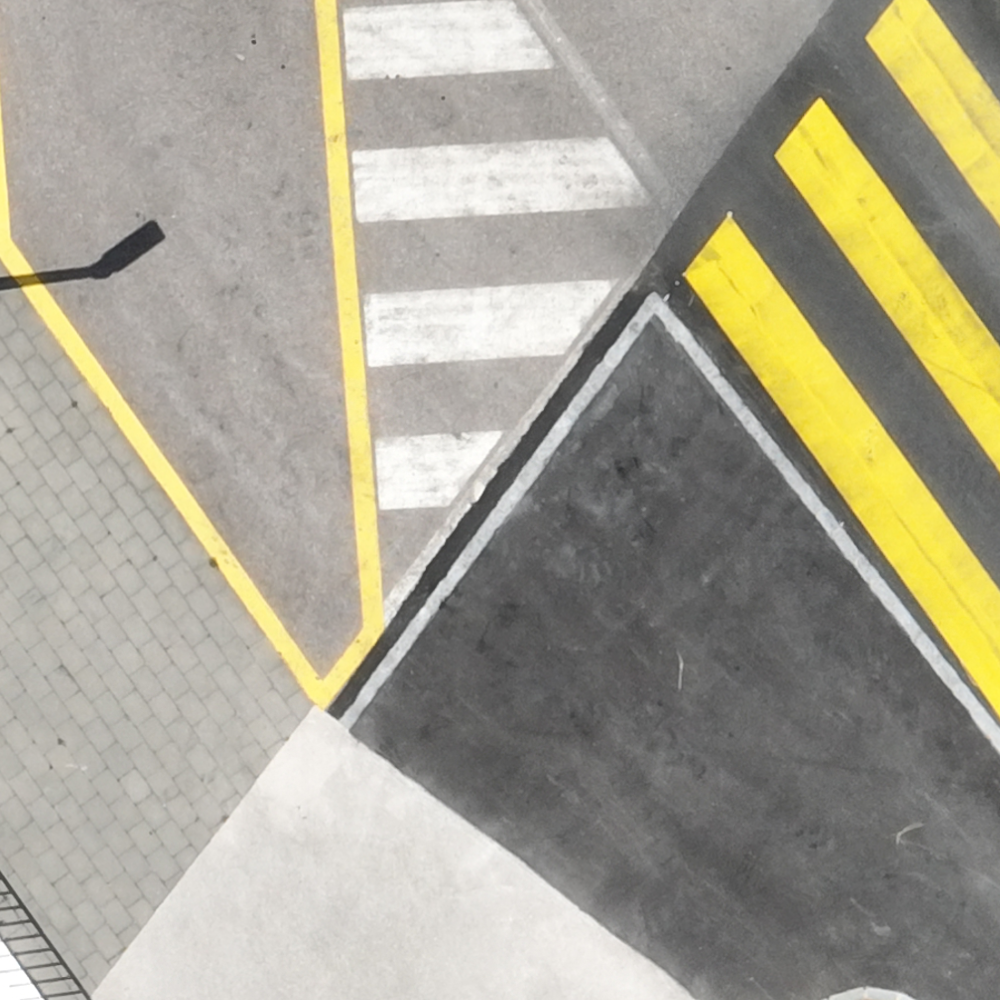}} \hfill
\subfloat[LR]
{\includegraphics[width=.45\columnwidth]{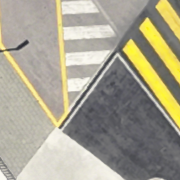}} \\
\subfloat[Pretrain (PSNR: 27.31, SSIM: 0.8557)]
{\includegraphics[width=.45\columnwidth]{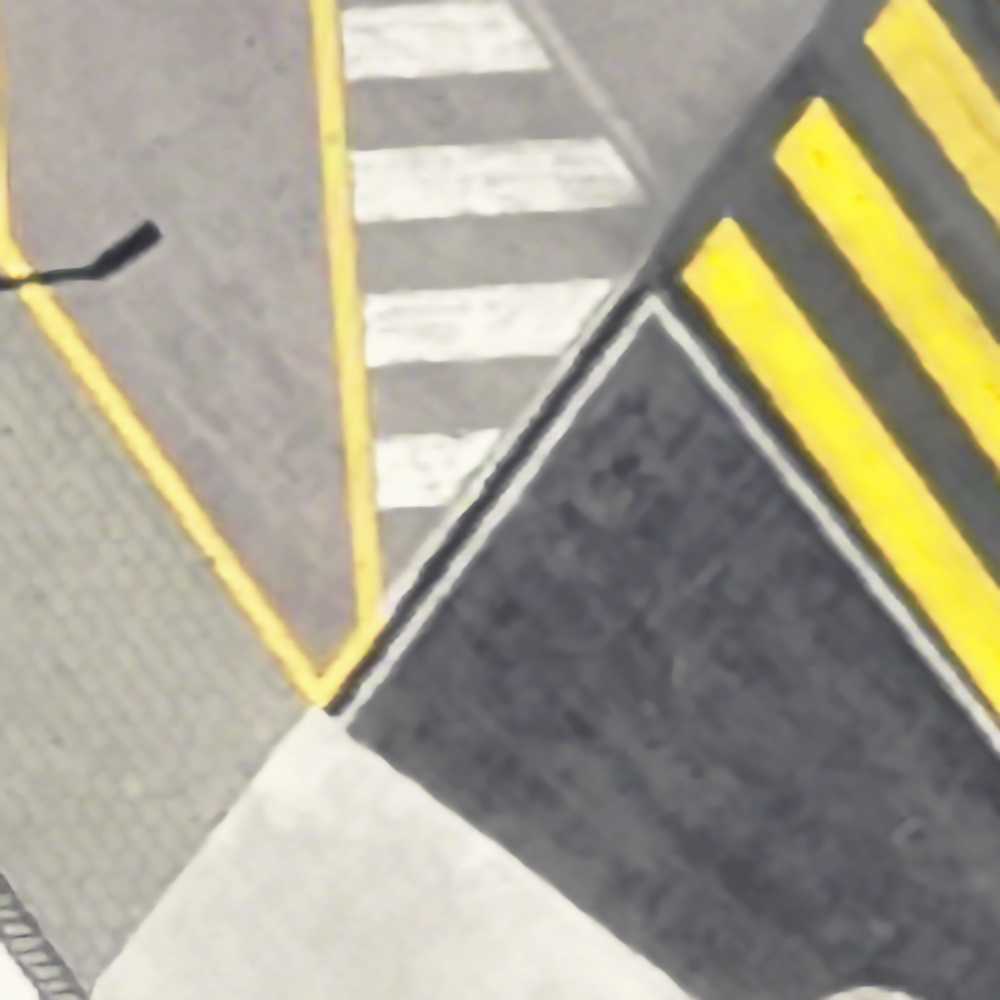}} \hfill
\subfloat[Fine-tune on all altitudes (PSNR: 27.75, SSIM: 0.8698)]
{\includegraphics[width=.45\columnwidth]{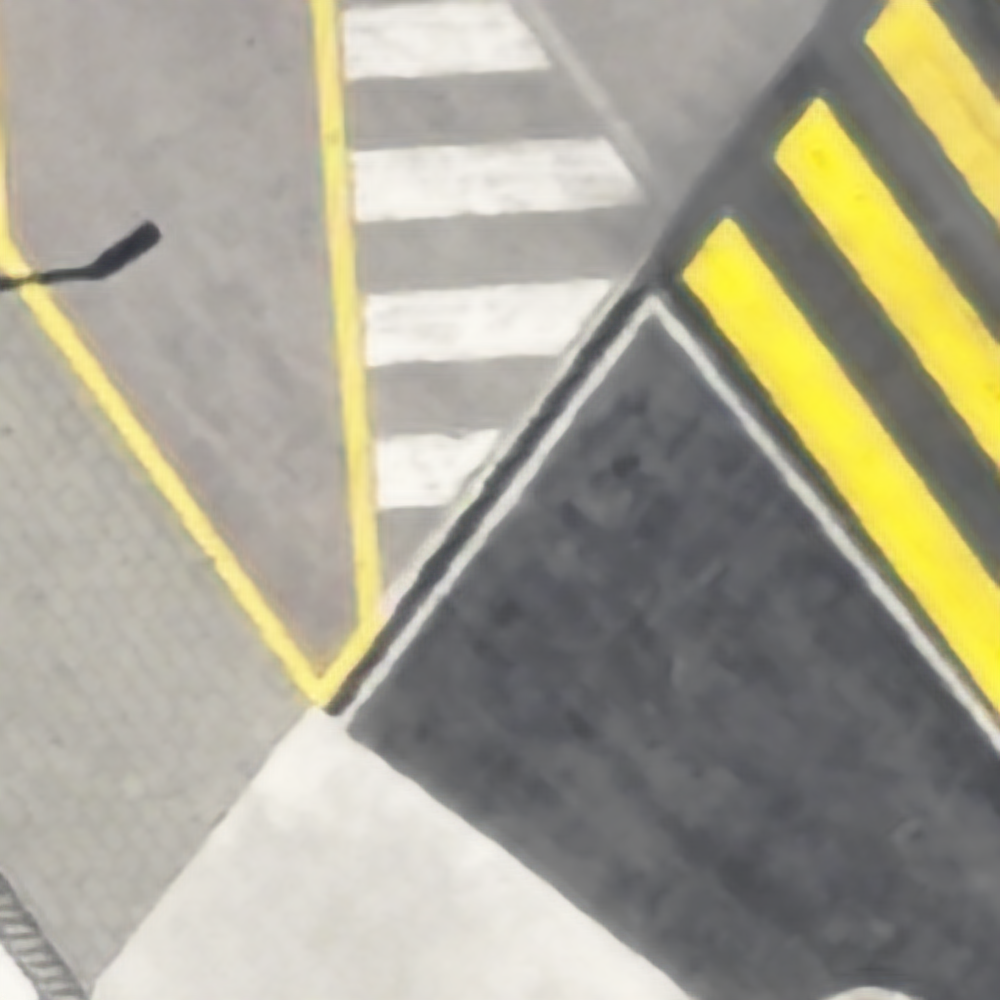}} \\
\subfloat[Fine-tune on 10m (PSNR: 27.42, SSIM: 0.8613)]
{\includegraphics[width=.45\columnwidth]{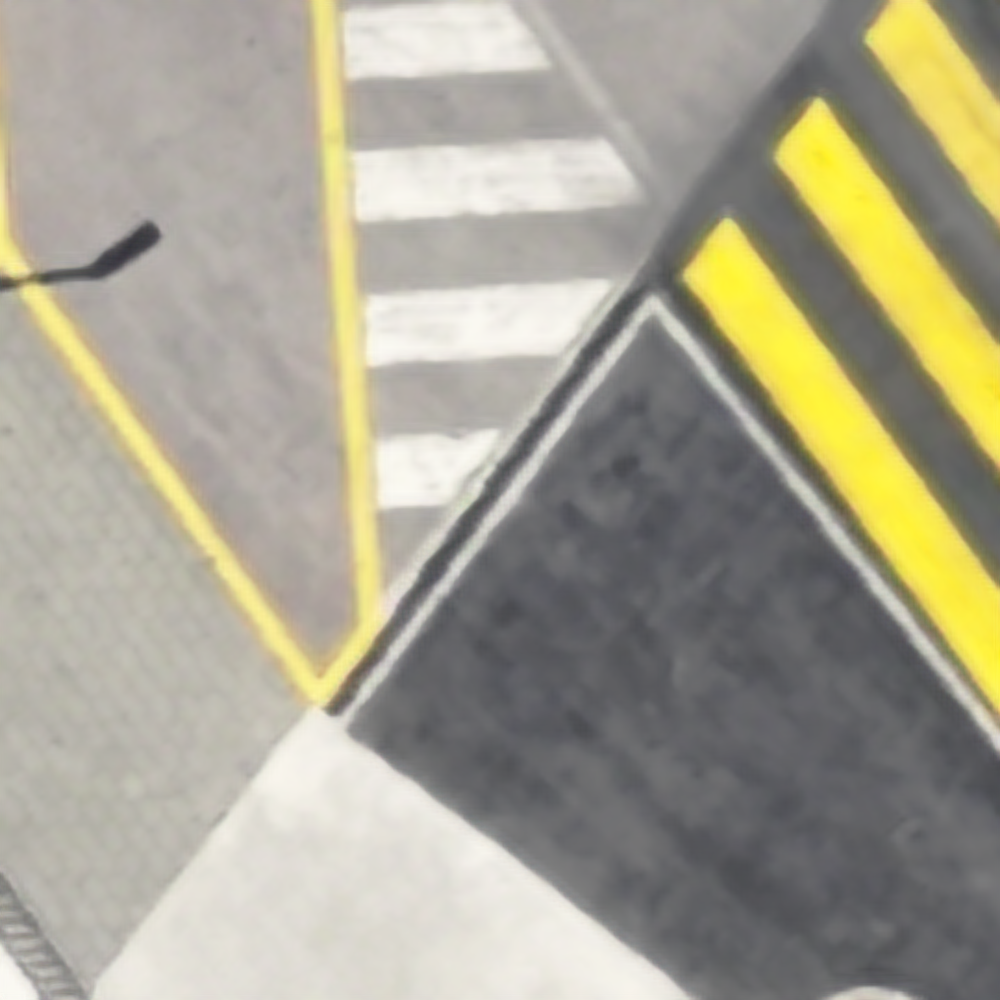}} \hfill
\subfloat[Fine-tune on 140m (PSNR: 27.80, SSIM: 0.8701)]
{\includegraphics[width=.45\columnwidth]{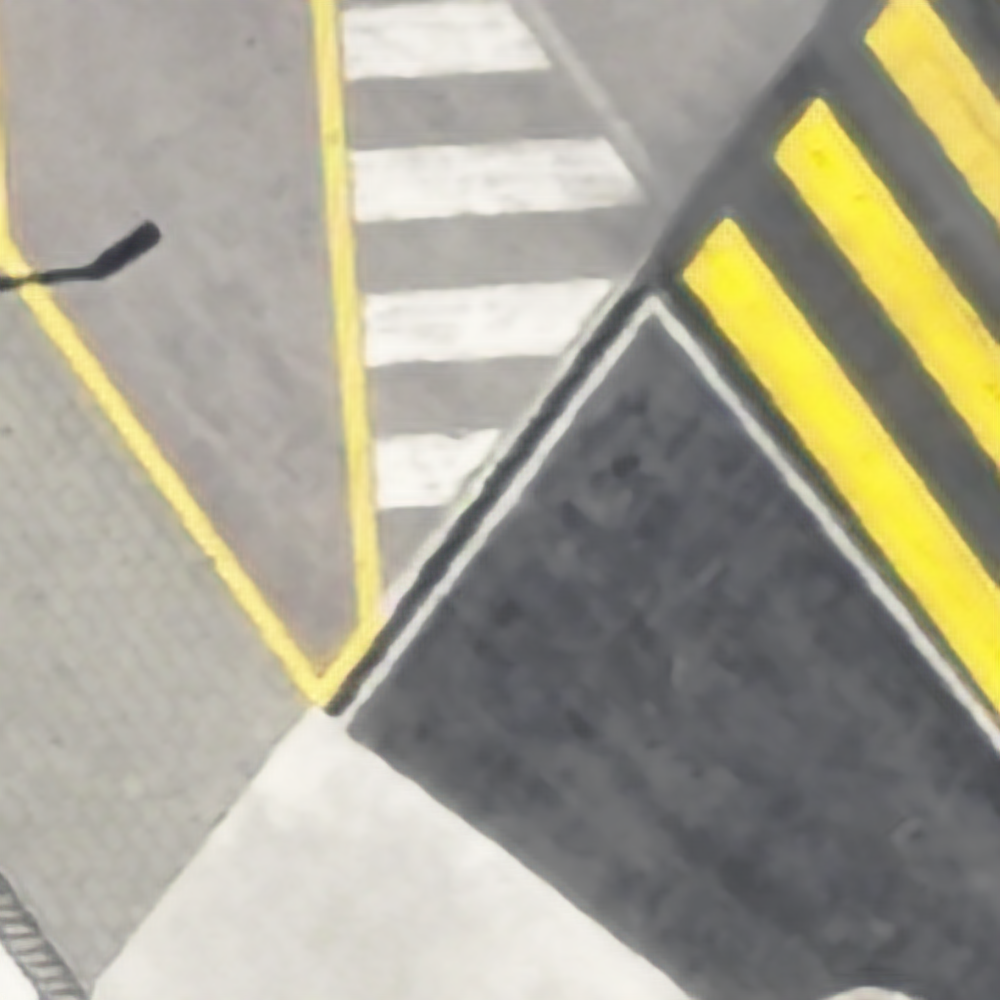}} \\
\caption{Test sample at 140m altitude}
\end{figure}

\section{Test samples of robust model for varying altitudes}
\label{sec: samples-robust}

\begin{figure}[tb]
\centering
\subfloat[HR]
{\includegraphics[width=.45\columnwidth]{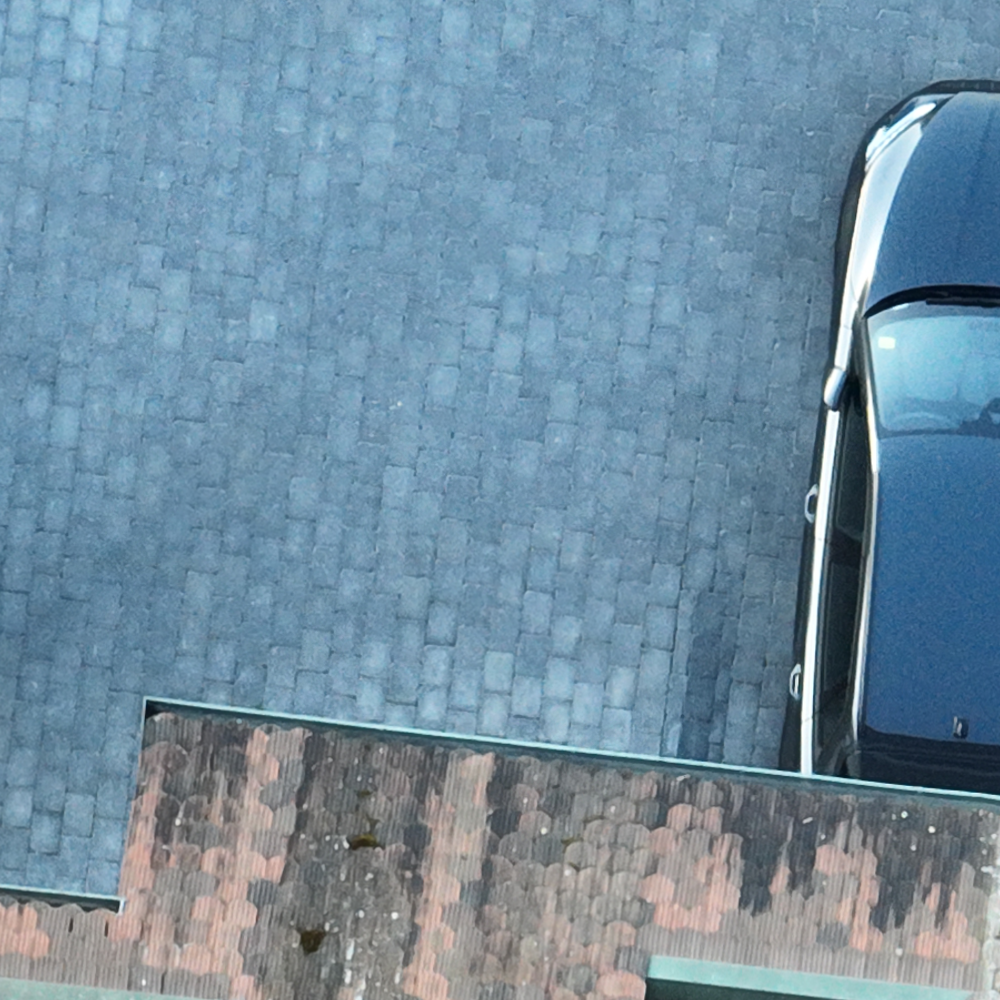}} \hfill
\subfloat[LR]
{\includegraphics[width=.45\columnwidth]{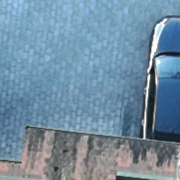}} \\
\subfloat[Without altitudes info (PSNR: 22.77, SSIM: 0.7349)]
{\includegraphics[width=.45\columnwidth]{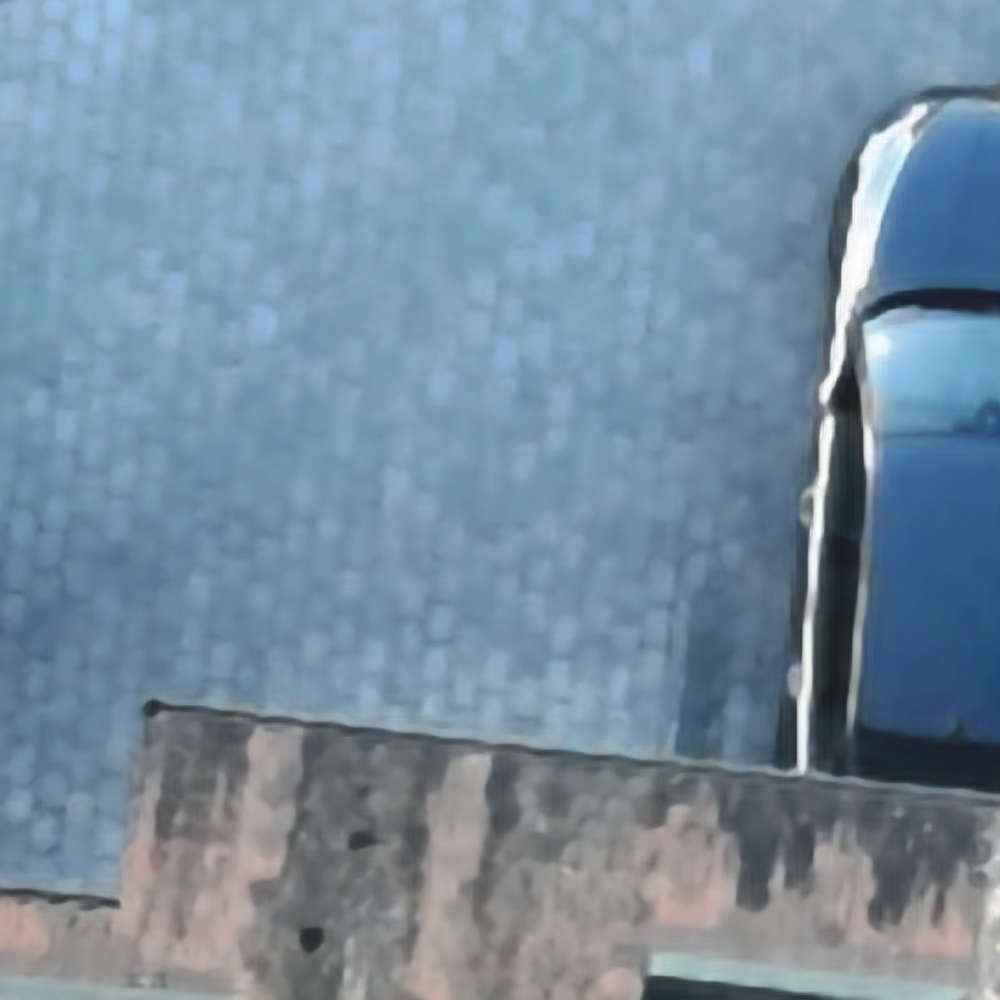}} \hfill
\subfloat[With altitude info (PSNR: 22.79, SSIM: 0.7390)]
{\includegraphics[width=.45\columnwidth]{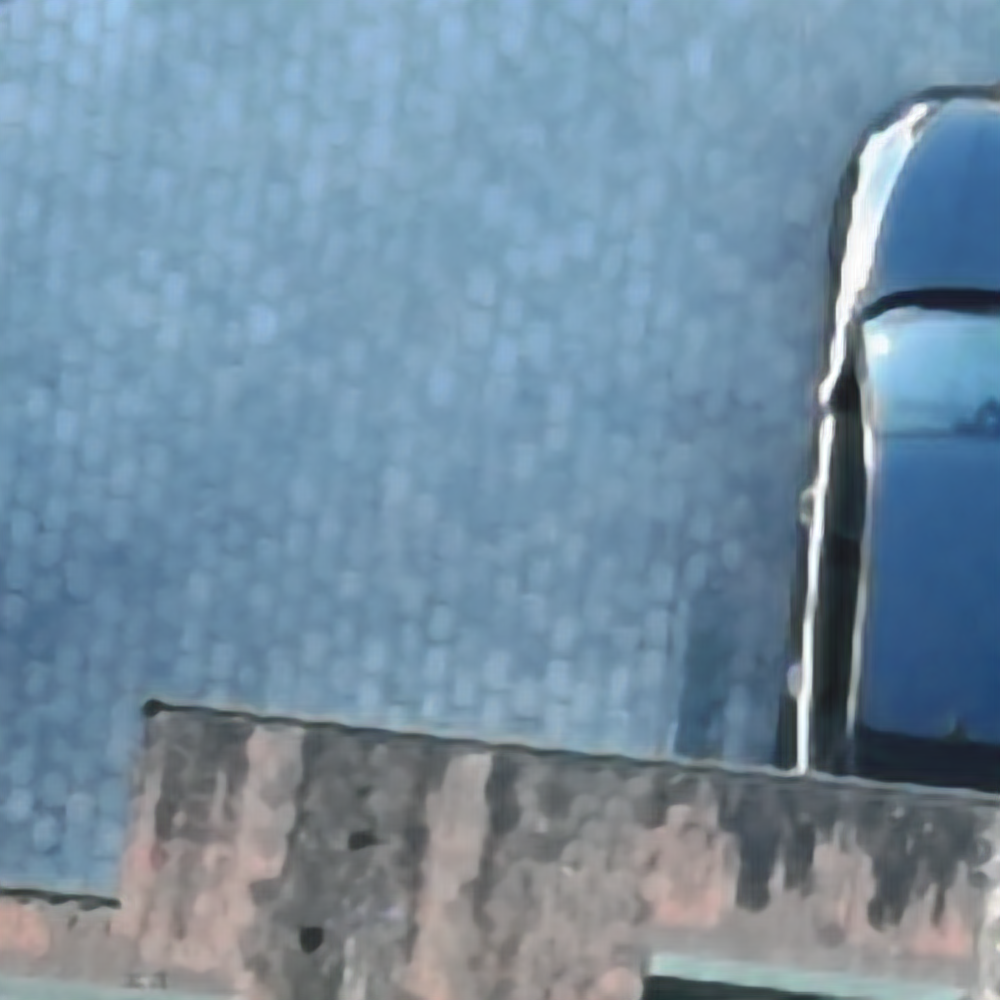}} \\
\caption{Test sample at 80m altitude}
\end{figure}
\backmatter
\cleardoublepage
\phantomsection
\printbibliography

\addcontentsline{toc}{chapter}{Bibliography}

\end{document}